\documentclass[acmsmall]{acmart}

\AtBeginDocument{%
  \providecommand\BibTeX{{%
    \normalfont B\kern-0.5em{\scshape i\kern-0.25em b}\kern-0.8em\TeX}}}

\usepackage{subfig}
\usepackage{multirow}
\usepackage{makecell}
\usepackage{mathrsfs}
\usepackage{ulem}
\usepackage{threeparttable}
\usepackage{color}
\definecolor{green}{RGB}{19,138,17}
\definecolor{yellow}{RGB}{248,148,6}
\usepackage{hyperref}

\begin{document}

\title{A Survey of Deep Active Learning}

\author{Pengzhen Ren}
\authornote{Both authors contributed equally to this research.}
\email{pzhren@foxmail.com}
\orcid{1234-5678-9012}
\author{Yun Xiao}
\authornotemark[1]
\email{yxiao@nwu.edu.cn}
\affiliation{%
  \institution{Northwest University}
}

\author{Xiaojun Chang}
\email{cxj273@gmail.com}
\affiliation{%
 \institution{RMIT University}
}

\author{Po-Yao Huang}
\affiliation{%
	\institution{Carnegie Mellon University}
}

\author{Zhihui Li}
\authornote{Corresponding author.}
\affiliation{%
	\institution{Qilu University of Technology (Shandong Academy of Sciences)}
}

\author{Brij B. Gupta}
\affiliation{
    \institution{National Institute of Technology Kurukshetra, India}
}

\author{Xiaojiang Chen}
\author{Xin Wang}
\affiliation{%
	\institution{Northwest University}
}

\renewcommand{\shortauthors}{Ren and Chang, et al.}

\begin{abstract}
Active learning (AL) attempts to maximize a model's performance gain while annotating the fewest samples possible. Deep learning (DL) is greedy for data and requires a large amount of data supply to optimize a massive number of parameters if the model is to learn how to extract high-quality features. In recent years, due to the rapid development of internet technology, we have entered an era of information abundance characterized by massive amounts of available data. As a result, DL has attracted significant attention from researchers and has been rapidly developed. Compared with DL, however, researchers have a relatively low interest in AL. This is mainly because before the rise of DL, traditional machine learning requires relatively few labeled samples, meaning that early AL is rarely according the value it deserves. Although DL has made breakthroughs in various fields, most of this success is due to a large number of publicly available annotated datasets. However, the acquisition of a large number of high-quality annotated datasets consumes a lot of manpower, making it unfeasible in fields that require high levels of expertise (such as speech recognition, information extraction, medical images, etc.). Therefore, AL is gradually coming to receive the attention it is due.

It is therefore natural to investigate whether AL can be used to reduce the cost of sample annotation while retaining the powerful learning capabilities of DL. As a result of such investigations, deep active learning (DeepAL) has emerged. Although research on this topic is quite abundant, there has not yet been a comprehensive survey of DeepAL-related works; accordingly, this article aims to fill this gap. We provide a formal classification method for the existing work, along with a comprehensive and systematic overview. In addition, we also analyze and summarize the development of DeepAL from an application perspective. Finally, we discuss the confusion and problems associated with DeepAL and provide some possible development directions.
\end{abstract}

\ccsdesc[500]{Computing methodologies ~Machine learning algorithms}

\keywords{Deep Learning, Active Learning, Deep Active Learning.}

\maketitle

\section{Introduction}

Both deep learning (DL) and active learning (AL) are a subfield of machine learning. DL is also called representation learning \cite{deng2014tutorial}. It originates from the study of artificial neural networks and realizes the automatic extraction of data features. DL has strong learning capabilities due to its complex structure, but this also means that DL requires a large number of labeled samples to complete the corresponding training. With the release of a large number of large-scale data sets with annotations and the continuous improvement of computer computing power, DL-related research has ushered in large development opportunities. Compared with traditional machine learning algorithms, DL has an absolute advantage in performance in most application areas. 
AL focuses on the study of data sets, and it is also known as query learning \cite{Settles2009ActiveLearningLiteratureSurvey}. AL assumes that different samples in the same data set have different values for the update of the current model, and tries to select the samples with the highest value to construct the training set. Then, the corresponding learning task is completed with the smallest annotation cost. Both DL and AL have important applications in the machine learning community. Due to their excellent characteristics, they have attracted widespread research interest in recent years. More specifically, DL has achieved unprecedented breakthroughs in various challenging tasks; however, this is largely due to the publication of massive labeled datasets \cite{Bengio2006GreedyLayerWise, Krizhevsky2012ImageNetClassificationwithDeepConvolutionalNeuralNetworks}. Therefore, DL is limited by the high cost of sample labeling in some professional fields that require rich knowledge. In comparison, an effective AL algorithm can theoretically achieve exponential acceleration in labeling efficiency \cite{Balcan2009Agnosticactivelearning}. This large potential saving in labeling costs is a fascinating development. However, the classic AL algorithm also finds it difficult to handle high-dimensional data \cite{2001Activelearningtheory}. Therefore, the combination of DL and AL, referred to as DeepAL, is expected to achieve superior results. DeepAL has been widely utilized in various fields, including image recognition \cite{Du2019BuildinganActivePalmprintRecognitionSystem, DBLP:conf/icml/GalIG17, DBLP:conf/cvpr/GudovskiyHYT20, Huang2020Cost}, text classification \cite{Zhang2016ActiveDiscriminativeText, schroder2020survey}, visual question answering \cite{Lin2017ActiveLearningVisual} and object detection \cite{Qu2020DeepActiveLearningforRemoteSensingObjectDetection, Aghdam2019Active, Feng2019Deep}, etc. Although a rich variety of related work has been published, DeepAL still lacks a unified classification framework. To fill this gap, in this article, we will provide a comprehensive overview of the existing DeepAL related work \footnote{We search about 270 related papers on \href{https://dblp.uni-trier.de/}{DBLP} using "deep active learning" as the keyword. We review the relevance of these papers to DeepAL one by one, eliminate irrelevant (just containing a few keywords) or information missing papers, and manually add some papers that do not contain these keywords but use DeepAL-related methods or relate to our current discussion. Finally, the survey references are constructed. The latest paper is updated to November 2020. The references include 103 conference papers, 153 journal papers, 3 books \cite{sanderson2008biometric, 2001Activelearningtheory, 2009Facilitylocationconcepts}, 1 research report \cite{Settles2009ActiveLearningLiteratureSurvey}, and 1 dissertation \cite{Zhu2005SemiSupervisedLearningwithGraphs}. There are 28 unpublished papers.}, along with a formal classification method. The contributions of this survey are summarized as follows:
\begin{itemize}
    \item As far as we know, this is the first comprehensive review work in the field of deep active learning.
    \item We analyze the challenges of combining active learning and deep learning, and systematically summarize and categorize existing DeepAL-related work for these challenges.
    \item We conduct a comprehensive and detailed analysis of DeepAL-related applications in various fields and future directions.
\end{itemize}

Next, we first briefly review the development status of DL and AL in their respective fields. Subsequently, in Section \ref{sec:The necessity and challenge of combining DL and AL}, the necessity and challenges of combining DL and AL are explicated. In Section \ref{sec: Deep Active Learning}, we conduct a comprehensive and systematic summary and discussion of the various strategies used in DeepAL. In Section \ref{sec: Various applications of DeepAL}, we review various applications of DeepAL in detail. In Section \ref{sec: Discussion and future directions}, we conduct a comprehensive discussion on the future direction of DeepAL. Finally, in Section \ref{sec: Summary and conclusions}, we make a summary and conclusion of this survey.

\subsection{Deep Learning}
\begin{figure}[!tp] 
	\centering 
	\subfloat[Structure diagram of convolutional neural network.] 
 	{\centering 
 		\includegraphics[width = 0.45\textwidth]{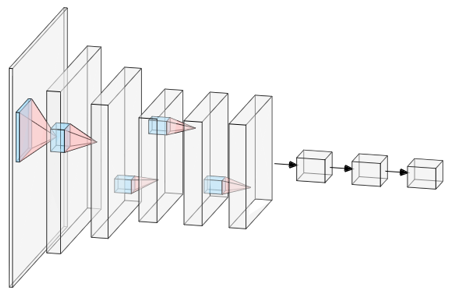}
 		\label{fig:DL}
 	}\hspace{5mm}
	\subfloat[The pool-based active learning cycle.] 
 	{\centering 
 		\includegraphics[width = 0.45\textwidth]{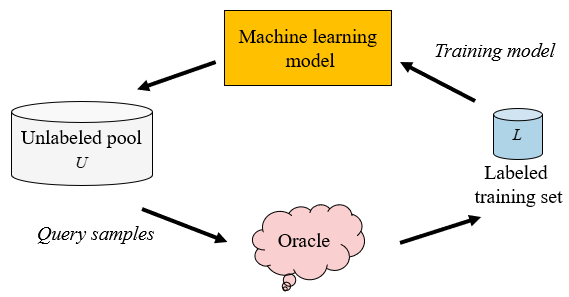}
 		\label{fig:AL}
 	}\\
	\subfloat[A typical example of deep active learning.] 
 	{\centering 
 		\includegraphics[width = 0.95\textwidth]{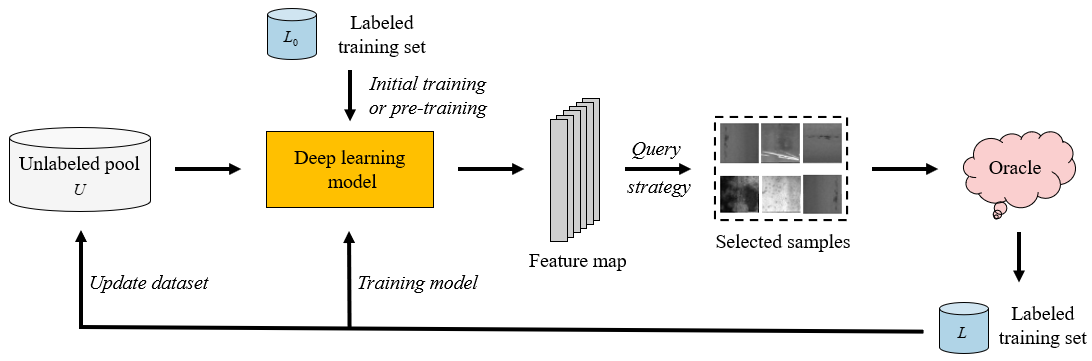}
 		\label{fig:DeepAL}
 	}
	\caption{Comparison of typical architectures of DL, AL, and DeepAL. (a) A common DL model: Convolutional Neural Network. (b)The pool-based AL cycle: Use the query strategy to query the sample in the unlabeled pool $U$ and hand it over to the oracle for labeling, then add the queried sample to the labeled training dataset $L$ and train, and then use the newly learned knowledge for the next round of querying. Repeat this process until the label budget is exhausted or the pre-defined termination conditions are reached. (c) A typical example of DeepAL: The parameters $\theta$ of the DL model are initialized or pre-trained on the labeled training set $L_0$, and the samples of the unlabeled pool $U$ are used to extract features through the DL model. Select samples based on the corresponding query strategy, and query the label in querying to form a new label training set $L$, then train the DL model on $L$, and update $U$ at the same time. Repeat this process until the label budget is exhausted or the pre-defined termination conditions are reached (see Section \ref{sec: DeepAL Stopping Strategy} for stopping strategy details).} 
	\label{fig:AL_DL_DAL} 
\end{figure}

DL attempts to build appropriate models by simulating the structure of the human brain. The McCulloch-Pitts (MCP) model proposed in 1943 by \cite{Fitch1944McCulloch} is regarded as the beginning of modern DL. Subsequently, in 1986, \cite{Rumelhart1986LearningRepresentationsbyBackPropagatingErrors} introduced backpropagation into the optimization of neural networks, which laid the foundation for the subsequent rapid development of DL. In the same year, Recurrent Neural Networks (RNNs) \cite{Jordan1986SerialOrderaParallelDistributedProcessingApproach} were first proposed. In 1998, the LeNet \cite{lecun1998gradient} network made its first appearance, representing one of the earliest uses of deep neural networks (DNN). However, these pioneering early works were limited by the computing resources available at the time and did not receive as much attention and investigation as they should have \cite{Lecun2015DeepLearning}. In 2006, Deep Belief Networks (DBNs) \cite{Hinton2006AFastLearningAlgorithmforDeepBeliefNets} were proposed and used to explore a deeper range of networks, which prompted the name of neural networks as DL. 
AlexNet \cite{Krizhevsky2012ImageNetClassificationwithDeepConvolutionalNeuralNetworks} is considered the first CNN deep learning model, which greatly improves the image classification results on large-scale data sets (such as ImageNet). In the ImageNet Large-Scale Visual Recognition Challenge (ILSVRC)-2012 competition \cite{deng2012large}, the AlexNet \cite{Krizhevsky2012ImageNetClassificationwithDeepConvolutionalNeuralNetworks} won the championship in the top-5 test error rate by nearly 10\% ahead of the second place. AlexNet uses the ReLUs (Rectified Linear Units) \cite{nair2010rectified} activation function to effectively suppress the gradient disappearance problem, while the use of multiple GPUs greatly improves the training speed of the model. 
Subsequently, DL began to win championships in various competitions and demonstrated very competitive results in many fields, such as visual data processing, natural language processing, speech processing, and many other well-known applications \cite{yan2017efficient, yan2015deep}. From the perspective of automation, the emergence of DL has transformed the manual design of features \cite{Dalal2005HistogramsofOrientedGradientsforHumanDetection, Lowe1999ObjectRecognitionfromLocalScaleInvariantFeatures} in machine learning to facilitate automatic extraction \cite{Simonyan2015VeryDeepConvolutionalNetworksforLargeScaleImageRecognition, He2016DeepResidualLearningforImageRecognition}. It is precisely because of this powerful automatic feature extraction capability that DL has demonstrated such unprecedented advantages in many fields. 
After decades of development, the research work related to DL is very rich. In Fig.\ref{fig:DL}, we present a standard deep learning model example: convolutional neural network (CNN) \cite{1986Learninginternalrepresentations, LeCun1989BackpropagationAppliedHandwritten}. Based on this approach, similar CNNs are applied to various image processing tasks. In addition, RNNs and GANs (Generative Adversarial Networks) \cite{Salvaris2018GenerativeAdversarialNetworks} are also widely utilized. Beginning in 2017, DL gradually shifted from the initial feature extraction automation to the automation of model architecture design \cite{Zoph2017NeuralArchitectureSearchwithReinforcementLearning, Ren2020AComprehensiveSurveyofNeuralArchitectureSearchChallengesandSolutions, DBLP:conf/iclr/BakerGNR17}; however, this still has a long way to go.

Thanks to the publication of a large number of existing annotation datasets \cite{Bengio2006GreedyLayerWise, Krizhevsky2012ImageNetClassificationwithDeepConvolutionalNeuralNetworks}, in recent years, DL has made breakthroughs in various fields including machine translation \cite{Aharoni2019Massively, Tan2019MultilingualNeuralMachineTranslationwithKnowledgeDistillation, Wang2019LearningDeepTransformerModelsforMachineTranslation, DBLP:conf/iclr/BauBSDDG19}, speech recognition \cite{DBLP:conf/interspeech/ParkCZCZCL19, Nassif2019SpeechRecognitionUsingDeepNeuralNetworksaSystematicReview, Qin2019ImperceptibleRobustandTargetedAdversarialExamplesforAutomaticSpeechRecognition, Schneider2019Wav2vecUnsupervisedPreTrainingforSpeechRecognition}, and image classification \cite{DBLP:conf/cvpr/HeZ0ZXL19, Mateen2019FundusImageClassificationUsingVGG19ArchitecturewithPCAandSVD, Paoletti2019CapsuleNetworksforHyperspectralImageClassification, Yalniz2019BillionScaleSemiSupervisedLearningforImageClassification}. However, this comes at the cost of a large number of manually labeled datasets, and DL has a strong greedy attribute to the data. While, in the real world, obtaining a large number of unlabeled datasets is relatively simple, the manual labeling of datasets comes at a high cost; this is particularly true for those fields where labeling requires a high degree of professional knowledge \cite{Hoi2006Batchmodeactive, Smith2018Lessismore}. For example, the labeling and description of lung lesion images of COVID-19 patients requires experienced clinicians to complete, and it is clearly impractical to demand that such professionals complete a large amount of medical image labeling. Similar fields also include speech recognition \cite{Zhu2005SemiSupervisedLearningwithGraphs, Abdelwahab2019Active}, medical imaging \cite{Hoi2006Batchmodeactive, nam2019development, lee2018a, Yang2017Suggestive}, recommender
systems \cite{Adomavicius2005nextgenerationrecommender, Cheng2019Deep}, information extraction \cite{Bhattacharjee2017Active}, satellite remote sensing \cite{Liu2017Active} and robotics \cite{Calinon2007LearningRepresentingGeneralizing, Andersson2017Deep, Takahashi2017Dynamic, Zhou2019Active, burka2017much}, machine translation \cite{bloodgood2014bucking,platanios2019competence} and text classification \cite{Zhang2016ActiveDiscriminativeText, schroder2020survey}, etc. Therefore, a way of maximizing the performance gain of the model when annotating a small number of samples is urgently required. 

\subsection{Active Learning}
AL is just such a method dedicated to studying how to obtain as many performance gains as possible by labeling as few samples as possible. More specifically, it aims to select the most useful samples from the unlabeled dataset and hand it over to the oracle (e.g., human annotator) for labeling, to reduce the cost of labeling as much as possible while still maintaining performance. AL approaches can be divided into membership query synthesis \cite{Angluin1988QueriesConceptLearning, king2004functional}, stream-based selective sampling \cite{dagan1995committee-based, krishnamurthy2002algorithms} and pool-based \cite{Lewis1994sequentialalgorithmtraining} AL from application scenarios \cite{Settles2009ActiveLearningLiteratureSurvey}. Membership query synthesis means that the learner can request to query the label of any unlabeled sample in the input space, including the sample generated by the learner. Moreover, the key difference between stream-based selective sampling and pool-based sampling is that the former makes an independent judgment on whether each sample in the data stream needs to query the labels of unlabeled samples, while the latter chooses the best query sample based on the evaluation and ranking of the entire dataset. Related research on stream-based selective sampling is mainly aimed at the application scenarios of small mobile devices that require timeliness, because these small devices often have limited storage and computing capabilities. The more common pool-based sampling strategy in the paper related to AL research is more suitable for large devices with sufficient computing and storage resources. In Fig.\ref{fig:AL}, we illustrate the framework diagram of the pool-based active learning cycle. In the initial state, we can randomly select one or more samples from the unlabeled pool $U$, give this sample to the oracle query label to get the labeled dataset $L$, and then train the model on $L$ using supervised learning. Next, we use this new knowledge to select the next sample to be queried, add the newly queried sample to $L$, and then conduct training. This process is repeated until the label budget is exhausted or the pre-defined termination conditions are reached (see Section \ref{sec: DeepAL Stopping Strategy} for stopping strategy details).

It is different from DL by using manual or automatic methods to design models with high-performance feature extraction capabilities. AL starts with datasets, primarily through the design of elaborate query rules to select the best samples from unlabeled datasets and query their labels, in an attempt to reduce the labeling cost to the greatest extent possible. Therefore, the design of query rules is crucial to the performance of AL methods. Related research is also quite rich. For example, in a given set of unlabeled datasets, the main query strategies include the uncertainty-based approach \cite{Lewis1994sequentialalgorithmtraining, Joshi2009Multiclassactive, Ranganathan2017Deep, Tong2002Supportvectormachine, Seung1992Querycommittee,DBLP:conf/cvpr/BeluchGNK18}, diversity-based approach \cite{Bilgic2009LinkbasedActive, Guo2010ActiveInstanceSampling, Nguyen2004Activelearningusing, DBLP:conf/icml/GalIG17} and expected model change \cite{Freytag2014SelectingInfluentialExamples, Roy2001OptimalActiveLearning, Settles2007MultipleInstanceActive}. In addition, many works have also studied hybrid query strategies \cite{Shui2019DeepActiveLearning, Yin2017DeepSimilarityBased, Ash2019DeepBatchActive, Zhdanov2019Diverseminibatch}, taking into account the uncertainty and diversity of query samples, and attempting to find a balance between these two strategies. Because separate sampling based on uncertainty often results in sampling bias \cite{Dasgupta2011Twofacesactive, DBLP:conf/naacl/BloodgoodS09}, the currently selected sample is not representative of the distribution of unlabeled datasets. On the other hand, considering only strategies that promote diversity in sampling may lead to increased labeling costs, as may be a considerable number of samples with low information content will consequently be selected. More classic query strategies are examined in \cite{Settles2012Active}. Although there is a substantial body of existing AL-related research, AL still faces the problem of expanding to high-dimensional data (e.g., images, text, and video, etc.) \cite{2001Activelearningtheory}; thus, most AL works tend to concentrate on low-dimensional problems \cite{2001Activelearningtheory, DBLP:conf/icml/Hernandez-Lobato15b}. In addition, AL often queries high-value samples based on features extracted in advance and does not have the ability to extract features.

\section{The necessity and challenge of combining DL and AL} 
\label{sec:The necessity and challenge of combining DL and AL}
DL has a strong learning capability in the context of high-dimensional data processing and automatic feature extraction, while AL has significant potential to effectively reduce labeling costs. Therefore, an obvious approach is to combine DL and AL, as this will greatly expand their application potential. This combined approach, referred to as DeepAL, was proposed by considering the complementary advantages of the two methods, and researchers have high expectations for the results of studies in this field. However, although AL-related research on query strategy is quite rich, it is still quite difficult to apply this strategy directly to DL. This is mainly due to:

\begin{itemize}
\item \textbf{Model uncertainty in Deep Learning.} The query strategy based on uncertainty is an important direction of AL research. In classification tasks, although DL can use the softmax layer to obtain the probability distribution of the label, the facts show that they are too confident. The SR (Softmax Response) \cite{Wang2017CostEffectiveActive} of the final output is unreliable as a measure of confidence, and the performance of this method will thus be even worse than that of random sampling \cite{Wang2014new}.

\item \textbf{Insufficient data for labeled samples.} AL often relies on a small amount of labeled sample data to learn and update the model, while DL is often very greedy for data \cite{Hinton2012Improvingneuralnetworks}. The labeled training samples provided by the classic AL method thus insufficient to support the training of traditional DL. In addition, the one-by-one sample query method commonly used in AL is also not applicable in the DL context \cite{Zhdanov2019Diverseminibatch}.

\item \textbf{Processing pipeline inconsistency.} The processing pipelines of AL and DL are inconsistent. Most AL algorithms focus primarily on the training of classifiers, and the various query strategies utilized are largely based on fixed feature representations. In DL, however, feature learning and classifier training are jointly optimized. Only fine-tuning the DL models in the AL framework, or treating them as two separate problems, may thus cause divergent issues \cite{Wang2017CostEffectiveActive}.
\end{itemize}

To address the first problem, some researchers have applied Bayesian deep learning \cite{Gal2015BayesianConvolutionalNeural} to deal with the high-dimensional mini-batch samples with fewer queries in the AL context \cite{DBLP:conf/icml/GalIG17, Pop2018DeepEnsembleBayesian, BatchBALD2019, Tran2019BayesianGenerativeActive}, thereby effectively alleviating the problem of the DL model being too confident about the output results.
To solve the problem of insufficient labelled sample data, researchers have considered using generative networks for data augmentation \cite{Tran2019BayesianGenerativeActive} or assigning pseudo-labels to high-confidence samples to expand the labeled training set \cite{Wang2017CostEffectiveActive}. Some researchers have also used labeled and unlabeled datasets to combine supervised and semisupervised training across AL cycles \cite{Simeoni2019Rethinkingdeepactive, Hossain2019Active}. In addition, the empirical research in \cite{Sener2018ActiveLearningConvolutional} shows that the previous heuristic-based AL \cite{Settles2009ActiveLearningLiteratureSurvey} query strategy is invalid when it is applied to DL in batch settings; therefore, for the one-by-one query strategy in classic AL, many researchers focus on the improvement of the batch sample query strategy \cite{BatchBALD2019, Zhdanov2019Diverseminibatch, Gissin2018DiscriminativeActiveLearning, Ash2019DeepBatchActive}, taking both the amount of information and the diversity of batch samples into account.
Furthermore, to deal with the pipeline inconsistency problem, researchers have considered modifying the combined framework of AL and DL to make the proposed DeepAL model as general as possible, an approach that can be extended to various application fields. This is of great significance to the promotion of DeepAL. For example, \cite{Yoo2019LearningLossActive} embeds the idea of AL into DL and consequently proposes a task-independent architecture design.

\section{Deep Active Learning}
\label{sec: Deep Active Learning}
In this section, we will provide a comprehensive and systematic overview of DeepAL-related works. Fig.\ref{fig:DeepAL} illustrates a typical example of DeepAL model architecture. The parameters $\theta$ of the deep learning model are initialized or pre-trained on the labeled training set $L_0$, while the samples of the unlabeled pool $U$ are used to extract features through the deep learning model. The next steps are to select samples based on the corresponding query strategy, and query the label in the oracle to form a new label training set $L$, then train the deep learning model on $L$ and update $U$ at the same time. This process is repeated until the label budget is exhausted or the predefined termination conditions are reached (see Section \ref{sec: DeepAL Stopping Strategy} for stopping strategy details). From the DeepAL framework example in Fig.\ref{fig:DeepAL}, we can roughly divide the DeepAL framework into two parts: namely, the AL query strategy on the unlabeled dataset and the DL model training method. These will be discussed and summarized in the following Section \ref{sec: Query Strategy Optimization in DeepAL} and \ref{sec: Insufficient Data in DeepAL} respectively. Next, we will discuss the efforts made by DeepAL on the generalization of the model in Section \ref{sec: Common Framework DeepAL}. Finally, we briefly discuss the stopping strategy in DeepAL in Section \ref{sec: DeepAL Stopping Strategy}.

\subsection{Query Strategy Optimization in DeepAL}
\label{sec: Query Strategy Optimization in DeepAL}
In the pool-based method, we define $U^n=\{\mathcal{X},\mathcal{Y}\}$ as an unlabeled dataset with $n$ samples; here, $\mathcal{X}$ is the sample space, $\mathcal{Y}$ is the label space, and $P(x,y)$ is a potential distribution, where $x\in \mathcal{X},y\in \mathcal{Y}$. $L^m=\{X,Y\}$ is the current labeled training set with $m$ samples, where $\mathrm{x}\in X,\mathrm{y}\in Y$. Under the standard supervision environment of DeepAL, our main goal is to design a query strategy $Q$, $U^n\stackrel{Q}{\longrightarrow}L^m$, using the deep model $f\in \mathcal{F },f:\mathcal{X}\rightarrow\mathcal{Y}$. The optimization problem of DeepAL in a supervised environment can be expressed as follows:
\begin{equation}
\mathop{\arg\min}_{L^m\subseteq U^n, (\mathrm{x,y}) \in L^m, (x,y) \in U^n} \mathbb{E}_{(x,y)}[\ell(f(\mathrm{x}),\mathrm{y})],
\end{equation}
where $\ell(\cdot)\in \mathcal{R}^+$ is the given loss equation, and we expect that $m\ll n$. Our goal is to make $m$ as small as possible while ensuring a predetermined level of accuracy. Therefore, the query strategy $Q$ in DeepAL is crucial to reduce the labeling cost.
Next, we will conduct a comprehensive and systematic review of DeepAL's query strategy from the following five aspects.
\begin{itemize}
    \item \textit{Batch Mode DeepAL (BMDAL).} The batch-based query strategy is the foundation of DeepAL. The one-by-one sample query strategy in traditional AL is inefficient and not applicable to DeepAL, so it is replaced by batch-based query strategy.
    \item \textit{Uncertainty-based and Hybrid Query Strategies.} Uncertainty-based query strategy refers to the model based on sample uncertainty ranking to select the sample to be queried. The greater the uncertainty of the sample, the easier it is to be selected. However, this is likely to ignore the relationship between samples. Therefore, the method that considers multiple sample attributes is called the hybrid query strategy.
    \item \textit{Deep Bayesian Active Learning (DBAL).} Active learning based on Bayesian convolutional neural network \cite{Gal2015BayesianConvolutionalNeural} is called deep Bayesian active learning.
    \item \textit{Density-based Methods.} The density-based method is a query strategy that attempts to find a core subset \cite{phillips2016coresets} representing the distribution of the entire dataset from the perspective of the dataset to reduce the cost of annotation.
    \item \textit{Automated Design of DeepAL.} Automated design of DeepAL refers to a method that uses automated methods to design AL query strategies or DL models that have an important impact on DeepAL performance.
\end{itemize}
\subsubsection{Batch Mode DeepAL (BMDAL)}
\label{sec:Batch Mode DeepAL (BMDAL)}
The main difference between DeepAL and classical AL is that DeepAL uses batch-based sample querying. In traditional AL, most algorithms use a one-by-one query method, which leads to frequent training of the learning model but little change in the training data. The training set obtained by this query method is not only inefficient in the training of the DL model, but can also easily lead to overfitting. Therefore, it is necessary to investigate BMDAL in more depth. In the context of BMDAL, at each acquisition step, we score a batch of candidate unlabeled data samples $\mathcal{B}=\{x_1,x_2,...,x_b\}\subseteq U$ based on the acquisition function $a$ used  and the deep model $f_{\theta}(L)$ trained on $L$, to select a new batch of data samples $\mathcal{B}^*=\{x_1^ *,x_2^*,...,x_b^*\}$. This problem can be formulated as follows:
\begin{equation}
\mathcal{B}^*= \mathop{\arg\max}_{\mathcal{B}\subseteq U} a_{batch}(\mathcal{B},f_{\theta}(L)),
\end{equation}
where $L$ is labeled training set. In order to facilitate understanding, we also use $ D_{train}$ to represent the labeled training set. 
\begin{figure}[!tp] 
	\centering 
	\subfloat[Batch query strategy considering only the amount of information.] 
 	{\centering 
 		\includegraphics[width = 0.4\textwidth]{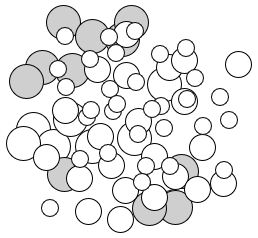}
 		\label{fig:One_by_one}
 	}\hspace{8mm}
	\subfloat[Batch query strategy considering both information volume and diversity.] 
 	{\centering 
 		\includegraphics[width = 0.4\textwidth]{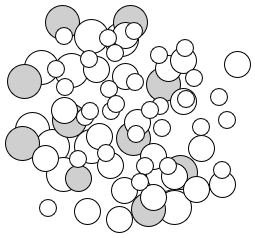}
 		\label{fig:Batch}
 	}
	\caption{A comparison diagram of two batch query strategies, one that only considers the amount of information and one that considers both the amount and diversity of information. The size of the dots indicates the amount of information in the samples, while the distance between the dots represents the similarity between the samples. The points shaded in gray indicate the sample points to be queried in a batch.} 
	\label{fig:OneBatch} 
\end{figure}

A naive approach would be to continuously query a batch of samples based on the one-by-one strategy. For example, \cite{Gal2016Dropout, Janz2017Actively} adopts the method of batch acquisition and chooses BALD (Bayesian Active Learning by Disagreement) \cite{Houlsby2011BayesianActiveLearning} to query top-$K$ samples with the highest scores. 
The acquisition function $ a_{BALD} $ of this idea is expressed as follows:
\begin{equation}
    \label{eq: BALD}
    \begin{aligned}
        &a_{\mathrm{BALD}}\left(\left\{x_{1}, \ldots, x_{b}\right\}, \mathcal{P}\left(\omega \mid D_{train}\right)\right)=\sum_{i=1}^{b} \mathbb{I}\left(y_{i} ; \omega \mid x_{i}, D_{train}\right),\\
        &\mathbb{I}\left(y ; \boldsymbol{\omega} \mid x, D_{train}\right)=\mathbb{H}\left(y \mid x, D_{train}\right)-\mathbb{E}_{\mathcal{P}\left(\boldsymbol{\omega} \mid D_{train}\right)}\left[\mathbb{H}\left(y \mid x, \boldsymbol{\omega}, D_{train}\right)\right],
    \end{aligned}
\end{equation}
where $ \mathbb{I}\left(y ; \boldsymbol{\omega}\mid x, D_{train}\right) $ used in BALD is to estimate the mutual information between model parameters and model predictions. The larger the mutual information value $\mathbb{I}(*)$, the higher the uncertainty of the sample. The condition of $\boldsymbol{\omega}$ on $D_{train}$ indicates that the model has been trained with $D_{train}$. And $\omega \sim \mathcal{P}\left(\omega \mid D_{train}\right)$ represents the model parameters of the current Bayesian model. $\mathbb{H(*)}$ represents the entropy of the model prediction. $\mathbb{E}[H(*)]$ is the expectation of the entropy of the model prediction over the posterior of the model parameters. Equation (\ref{eq: BALD}) considers each sample independently and selects samples to construct a batch query dataset in a one-by-one way.

Clearly, however, this method is not feasible, as it is very likely to choose a set of information-rich but similar samples. The information provided to the model by such similar samples is essentially the same, which not only wastes labeling resources, but also makes it difficult for the model to learn genuinely useful information. In addition, this query method that considers each sample independently also ignores the correlation between samples. This is likely to lead to local decisions that make the batch sample set of queries insufficiently optimized. Therefore, how to simultaneously consider the correlation between different query samples is the primary problem for BMDAL. To solve the above problems, BatchBALD \cite{BatchBALD2019} expands BALD, which considers the correlation between data points by estimating the joint mutual information between multiple data points and model parameters. The acquisition function of BatchBALD can be expressed as follows:
\begin{equation}
\begin{aligned}
    &a_{\text {BatchBALD }}\left(\left\{x_{1}, \ldots, x_{b}\right\}, \mathcal{P}\left(\omega \mid D_ {train }\right)\right)=\mathbb{I}\left(y_{1}, \ldots, y_{b} ; \omega \mid x_{1}, \ldots, x_{b}, D_ {train }\right),\\
    &\mathbb{I}\left(y_{1: b} ; \boldsymbol{\omega} \mid x_{1: b}, D_ {train }\right)=\mathbb{H}\left(y_{1: b} \mid x_{1: b}, D_ {train }\right)-\mathbb{E}_{\mathcal{P}\left(\boldsymbol{\omega} \mid D_ {train }\right)} \mathbb{H}\left(y_{1: b} \mid x_{1: b}, \boldsymbol{\omega}, D_ {train }\right),
\end{aligned}
\end{equation}
where $x_{1}, \ldots, x_{b}$ and $y_{1}, \ldots, y_{b}$ are represented by joint random variables $x_{1: b}$ and $y_{1: b}$ in a product probability space, and $\mathbb{I}\left(y_{1: b} ; \boldsymbol{\omega} \mid x_{1: b}, D_ {train }\right)$ denotes the mutual information between these two random variables. BatchBALD considers the correlation between different query samples by designing an explicit joint mutual information mechanism to obtain a better query batch sample set. 

The batch-based query strategy forms the basis of the combination of AL and DL, and related research on this topic is also very rich. We will provide a detailed overview and discussion of BMDAL query strategies in the following sections.

\subsubsection{Uncertainty-based and Hybrid Query Strategies}
\label{sec: Hybrid query strategy}
Because the uncertainty-based approach is simple in form and has low computational complexity, it is a very popular query strategy in AL. This query strategy is mainly used in certain shallow models (eg, SVM \cite{Tong2002Supportvectormachine} or KNN \cite{Jain2009Active}). This is mainly because the uncertainty of these models can be accurately obtained by traditional uncertainty sampling methods.
In uncertainty-based sampling, learners try to select the most uncertain samples to form a batch query set. For example, in the margin sampling \cite{DBLP:conf/ida/SchefferDW01}, margin $M$ is defined as the difference between the predicted highest probability and the predicted second highest probability of an sample as follows: 
$M=P\left(y_{1} \mid x\right)-P\left(y_{2} \mid x\right),$
where \(y_1\) and \(y_2\) are the first and second most probable labels predicted for the sample \(x\) under the current model. The smaller the margin $M$, the greater the uncertainty of the sample $x$. The AL algorithm selects the top-$K$ samples with the smallest margin $M$ as the batch query set by calculating the margin $M$ of all unlabeled samples.
Information entropy \cite{Settles2009ActiveLearningLiteratureSurvey} is also a commonly used uncertainty measurement standard. For a $k$-class task, the information entropy $\mathbb{E}(x)$ of sample $x$ can be defined as follows:
\begin{equation}
\label{eq: entropy}
\mathbb{E}(x)=-\sum_{i=1}^{k} P(y_i\mid x) \cdot \log \left(P(y_i\mid x)\right),
\end{equation}
where $P(y_i\mid x)$ is the probability that the current sample $x$ is predicted to be class $y_i$. The greater the entropy of the sample, the greater its uncertainty. Therefore, the top-$K$ samples with the largest information entropy should be selected. More query strategies based on uncertainty can be found in \cite{DBLP:books/crc/aggarwal14/AggarwalKGHY14}.

There are many DeepAL \cite{Ranganathan2017Deep, Asghar2016Deep, He2019Towards, Ostapuk2019ActiveLink} methods that directly utilize an uncertainty-based sampling strategy.
However, DFAL (DeepFool Active Learning) \cite{Ducoffe2018Adversarial} contends that these methods are easily fooled by adversarial examples; thus, it focuses on the study of examples near the decision boundary, and actively uses the information provided by these adversarial examples on the input spatial distribution in order to approximate their distance to the decision boundary. This adversarial query strategy can effectively improve the convergence speed of CNN training. 
Nevertheless, as analyzed in Section \ref{sec:Batch Mode DeepAL (BMDAL)}, this can easily lead to insufficient diversity of batch query samples (such that relevant knowledge regarding the data distribution is not fully utilized), which in turn leads to low or even invalid DL model training performance. 
A feasible strategy would thus be to use a hybrid query strategy in a batch query, taking into account both the information volume and diversity of samples in either an explicit or implicit manner.

The performance of early Batch Mode Active Learning (BMAL) \cite{Wang2016batchmodeactive, Joshi2010Multiclassbatch,Brinker2003Incorporatingdiversityactive, Xia2016CostSensitiveBatch, Nguyen2004Activelearningusing, DBLP:conf/cikm/TanYHD19} algorithms are often excessively reliant on the measurement of similarity between samples. In addition, these algorithms are often only good at exploitation (learners tend to focus only on samples near the current decision boundary, corresponding to high-information query strategies), meaning that the samples in the query batch sample set cannot represent the true data distribution of the feature space (due to the insufficient diversity of batch sample sets). To address this issue, Exploration-P \cite{Yin2017DeepSimilarityBased} uses a deep neural network to learn the feature representation of the samples, then explicitly calculates the similarity between the samples. At the same time, the processes of exploitation and exploration (in the early days of model training, learners used random sampling strategies for exploration purposes) are balanced to enable more accurate measurement of the similarity between samples. 
More specifically, Exploration-P uses the information entropy in Equation (\ref{eq: entropy}) to estimate the uncertainty of sample $x$ under the current model. The uncertainty of the selected sample set $S$ can be expressed as $E(S) = \sum_{x_i\in S}\mathbb{E}(x_i)$. Furthermore, to measure the redundancy between samples in the selected sample set $S$, Exploration-P uses $R(S)$ to represent the redundancy of selected sample set $S$:
\begin{equation}
    R(S) = \sum_{x_i\in S}\sum_{x_j\in S}Sim(x_i,x_j),\quad Sim(x_i,x_j) = f(x_i)\mathcal{M}f(x_j),
\end{equation}
where $f(x)$ represents the feature of sample $x$ extracted by deep learning model $f$, $Sim(x_i,x_j)$ measures the similarity between two samples, and $\mathcal{M}$ is a similarity matrix (when $\mathcal{M}$ is the identity matrix, the similarity of two samples is the product of their feature vectors. In addition, $\mathcal{M}$ can also be learned as a parameter of $f$). Therefore, the selected sample set $S$ is expected to have the largest uncertainty and the smallest redundancy. For this reason, Exploration-P considers these two strategies, and the final goal equation is defined as:
\begin{equation}
\mathrm{I}(S)=E(S)-\frac{\alpha}{|S|} R(S),
\end{equation}
where, $\alpha$ is used to balance the weight of the hybrid query strategies, uncertainty and redundancy.

Moreover, DMBAL (Diverse Mini-Batch Active Learning) \cite{Zhdanov2019Diverseminibatch} adds informativeness to the optimization goal of K-means by weight, and further presents an in-depth study of a hybrid query strategy that considers the sample information volume and diversity under the mini-batch sample query setting. DMBAL \cite{Zhdanov2019Diverseminibatch} can easily achieve expansion from the generalized linear model to DL; this not only increases the scalability of DMBAL \cite{Zhdanov2019Diverseminibatch} but also increases the diversity of active query samples in the mini-batch. Fig.\ref{fig:OneBatch} illustrates a schematic diagram of this idea. This hybrid query strategy is quite popular.  For example, WI-DL (Weighted Incremental Dictionary Learning) \cite{Liu2017Active} mainly considers the two stages of DBN. In the unsupervised feature learning stage, the key consideration is the representativeness of the data, while in the supervised fine-tuning stage, the uncertainty of the data is considered; these two indicators are then integrated, and finally optimized using the proposed weighted incremental dictionary learning algorithm.

\begin{figure}[!tp] 
	\centering 
	\includegraphics[width = 0.95\textwidth]{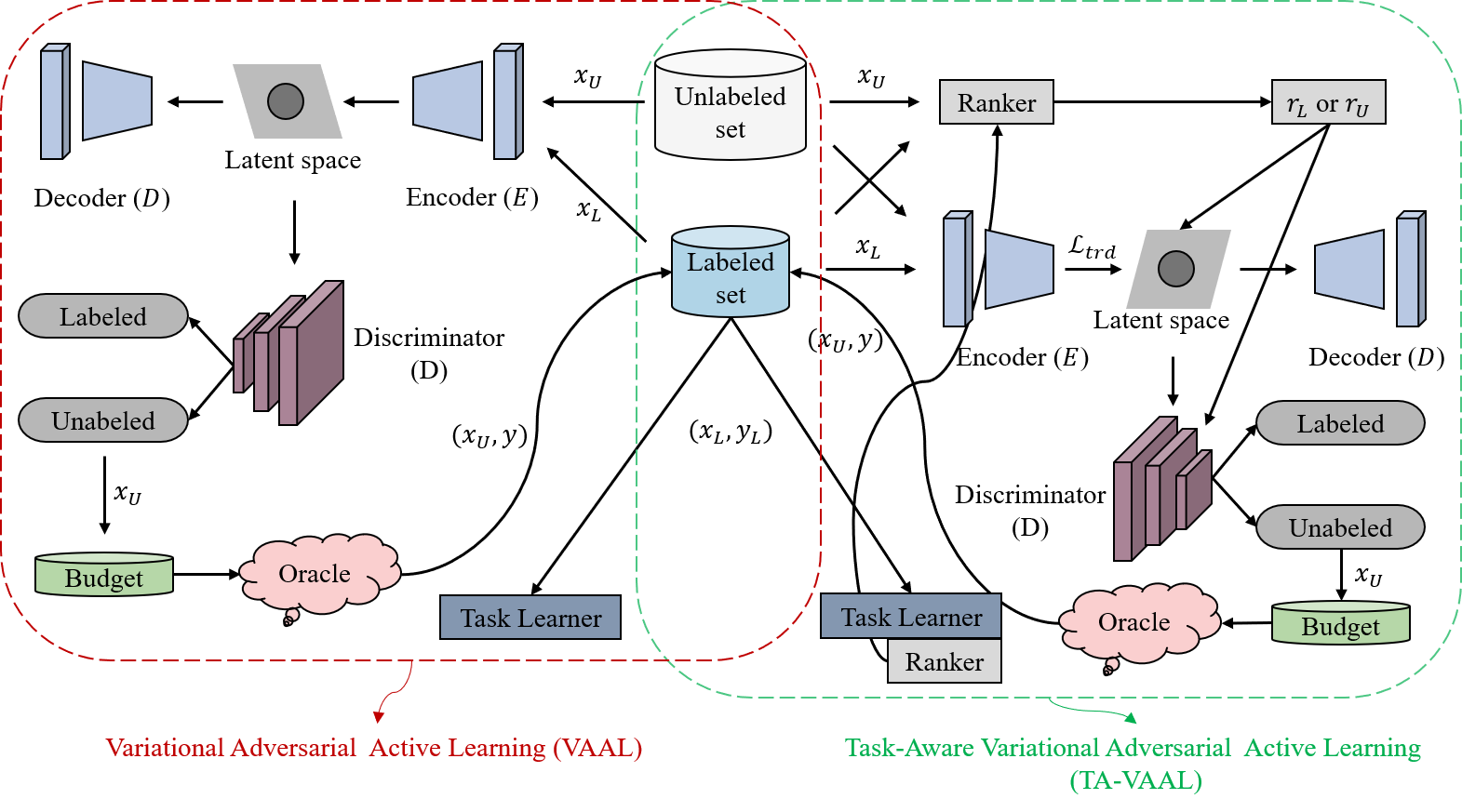}
	\caption{Structure comparison chart of VAAL \cite{sinha2019variational} and TA-VAAL \cite{Kim2020Task}. 1) VAAL uses labeled data and unlabeled data in a semi-supervised way to learn the latent representation space of the data, then selects the unlabeled data with the largest amount of information according to the latent space for labeling. 2) TA-VAAL expands VAAL and integrates the loss prediction module \cite{Yoo2019LearningLossActive} and RankCGAN \cite{Saquil2018Ranking} into VAAL in order to consider data distribution and model uncertainty simultaneously.} 
	\label{fig:TA-VAAL}
\end{figure}

Although the above improvements have resulted in a good performance, there is still a hidden danger that must be addressed: namely, that, diversity-based strategies are not appropriate for all datasets. More specifically, the richer the category content of the dataset, the larger the batch size, and the better the effect of diversity-based methods; by contrast, an uncertainty-based query strategy will perform better with smaller batch sizes and less rich content. These characteristics depend on the statistical characteristics of the dataset. The BMAL context, whether the data are unfamiliar and potentially unstructured, makes it impossible to determine which AL query strategy is more appropriate. In light of this, BADGE (Batch Active learning by Diverse Gradient Embeddings) \cite{Ash2019DeepBatchActive} samples point groups that are disparate and high magnitude when represented in a hallucinated gradient space, meaning that both the prediction uncertainty of the model and the diversity of the samples in a batch are considered simultaneously. Most importantly, BADGE can achieve an automatic balance between forecast uncertainty and sample diversity without the need for manual hyperparameter adjustments. 
Moreover, while BADGE \cite{Ash2019DeepBatchActive} considers this hybrid query strategy in an implicit way, WAAL (Wasserstein Adversarial Active Learning) \cite{Shui2019DeepActiveLearning} proposes a hybrid query strategy that explicitly balances uncertainty and diversity. In addition, WAAL \cite{Shui2019DeepActiveLearning} uses Wasserstein distance to model the interactive procedure in AL as a distribution matching problem, derives losses from it, and then decomposes WAAL \cite{Shui2019DeepActiveLearning} into two stages: DNN parameter optimization and query batch selection. 
TA-VAAL (Task-Aware Variational Adversarial Active Learning) \cite{Kim2020Task} also explores the balance of this hybrid query strategy. The assumption underpinning TA-VAAL is that the uncertainty-based method does not make good use of the overall data distribution, while the data distribution-based method often ignores the structure of the task. Consequently, TA-VAAL proposes to integrate the loss prediction module \cite{Yoo2019LearningLossActive} and the concept of RankCGAN \cite{Saquil2018Ranking} into VAAL (Variational Adversarial Active Learning) \cite{sinha2019variational}, enabling both the data distribution and the model uncertainty to be considered. TA-VAAL has achieved good performance on various balanced and unbalanced benchmark datasets. The structure diagram of TA-VAAL and VAAL is presented in Fig.\ref{fig:TA-VAAL}.

Notably, although the hybrid query strategy achieves superior performance, the uncertainty-based AL query strategy is more convenient to combine with the output of the softmax layer of DL. Thus, the query strategy based on uncertainty is still widely used.
\subsubsection{Deep Bayesian Active Learning (DBAL)}
 As noted in Section \ref{sec:The necessity and challenge of combining DL and AL}, which analyzes the challenge of combining DL and AL, the acquisition function based on uncertainty is an important research direction of many classic AL algorithms. Moreover, traditional DL methods rarely represent such model uncertainty.
 
 To solve the above problems, Deep Bayesian Active Learning appears. In the given input set $X$ and the output $Y$ belonging to class $c$, the probabilistic neural network model can be defined as $f(\mathrm{x};\theta)$, $p(\theta)$ is a prior on the parameter space $\theta$ (usually Gaussian), and the likelihood $p(\mathrm{y} = c|\mathrm{x},\theta)$ is usually given by $softma\mathrm{x}(f(\mathrm{x};\theta))$. Our goal is to obtain the posterior distribution over $\theta$, as follows:
 \begin{equation}
 p(\theta|X, Y)=\frac{p(Y | X, \theta) p(\theta)}{p(Y | X)}.
 \end{equation}
 For a given new data point $\mathrm{x}^*$, $\hat{\mathrm{y}}$ is predicted by:
 \begin{equation}
 p\left(\hat{\mathrm{y}} | \mathrm{x}^{*}, X, Y\right)=\int p\left(\hat{\mathrm{y}} | \mathrm{x}, \theta\right) p(\theta | X, Y) d \theta=\mathbb{E}_{\theta \sim p(\theta | X, Y)}[f(\mathrm{x} ; \theta)].
 \end{equation}

DBAL \cite{DBLP:conf/icml/GalIG17} combines BCNNs (Bayesian Convolutional Neural Networks) \cite{Gal2015BayesianConvolutionalNeural} with AL methods to adapt BALD \cite{Houlsby2011BayesianActiveLearning} to the deep learning environment, thereby developing a new AL framework for high-dimensional data. This approach adopts the above method to first perform Gaussian prior modeling on the weights of a CNN, and then uses variational inference to obtain the posterior distribution of network prediction. In addition, in practice, researchers often also use a powerful and low-cost MC-dropout (Monte-Carlo dropout) \cite{srivastava2014dropout:} stochastic regularization technique to obtain posterior samples, consequently attaining good performance on real-world datasets \cite{Leibig2017Leveraging, Kendall2015Bayesian}. Moreover, this regularization technique has been proven to be equivalent to variational inference \cite{Gal2016Dropout}. 
However, a core-set approach \cite{Sener2018ActiveLearningConvolutional} points out that DBAL \cite{DBLP:conf/icml/GalIG17} is unsuitable for large datasets due to the need for batch sampling. It should be noted here that while DBAL \cite{DBLP:conf/icml/GalIG17} allows the use of dropout in testing for better confidence estimation, the analysis presented in \cite{Gissin2018DiscriminativeActiveLearning} contends that the performance of this method is similar to the performance of using neural network SR \cite{Wang2017CostEffectiveActive} as uncertainty sampling, which requires vigilance. 
In addition, 
DEBAL (Deep Ensemble Bayesian Active Learning) \cite{Pop2018DeepEnsembleBayesian} argues that the pattern collapse phenomenon \cite{Srivastava2017VEEGAN} in the variational inference method leads to the overconfident prediction characteristic of the DBAL method. For this reason, DEBAL combines the expressive power of ensemble methods with MC-dropout to obtain better uncertainty in the absence of trading representativeness.
For its part, BatchBALD \cite{BatchBALD2019} opts to expand BALD \cite{Houlsby2011BayesianActiveLearning} to the batch query context; this approach no longer calculates the mutual information between a single sample and model parameters but rather recalculates the mutual information between the batch samples and the model parameters to jointly score the batch of samples. This enables BatchBALD to more accurately evaluate the joint mutual information. Inspired by the latest research on Bayesian core sets \cite{Huggins2016Coresets, Campbell2019Automated}, ACS-FW (Active Bayesian CoreSets with Frank-Wolfe optimization) \cite{Pinsler2019Bayesian} reconstructed the batch structure to optimize the sparse subset approximation of the log-posterior induced by the entire dataset. Using this similarity, ACS-FW then employs the Frank-Wolfe \cite{Frank1956algorithm} algorithm to enable effective Bayesian AL at scale, while its use of random projection has made it still more popular. Compared with other query strategies (e.g., maximizing the predictive entropy
(MAXENT) \cite{Sener2018ActiveLearningConvolutional, DBLP:conf/icml/GalIG17} and BALD \cite{Houlsby2011BayesianActiveLearning}), ACS-FW achieves better coverage across the entire data manifold. 
DPEs (Deep Probabilistic Ensembles) \cite{Chitta2018Large} introduces an expandable DPEs technology, which uses a regularized ensemble to approximate the deep BNN, and then evaluates the classification effect of these DPEs in a series of large-scale visual AL experiments.

ActiveLink (Deep Active Learning for Link Prediction in Knowledge Graphs) \cite{Ostapuk2019ActiveLink} is inspired by the latest advances in Bayesian deep learning \cite{Gal2016Dropout, Welling2011Bayesian}. Adopting the Bayesian view of the existing neural link predictors, it expands the uncertainty sampling method by using the basic structure of the knowledge graph, thereby creating a novel DeepAL method. ActiveLink further noted that although AL can sample efficiently, the model needs to be retrained from scratch for each iteration in the AL process, which is unacceptable in the DL model training context. A simple solution would be to use newly selected data to train the model incrementally, or to combine it with existing training data \cite{Shen2017DeepActiveLearning}; however, this would cause the model to be biased either towards a small amount of newly selected data or towards data selected early in the process. In order to solve this bias problem, ActiveLink adopts a principled and unbiased incremental training method based on meta-learning. More specifically, in each AL iteration, ActiveLink uses the newly selected samples to update the model parameters, then approximates the meta-objective of the model's future prediction by generalizing the model based on the samples selected in the previous iteration. This enables ActiveLink to strike a balance between the importance of the newly and previously selected data, and thereby to achieve an unbiased estimation of the model parameters.

In addition to the above-mentioned DBAL work, due to the lesser parameter of BNN and the uncertainty sampling strategy being similar to traditional AL, the research on DBAL is quite extensive, and there are many works related to this topic \cite{Siddhant2018Deep, Rottmann2018Deep, Yang2018Leveraging, Zeng2018Relevance, Gudur2019Activeharnet, MartinezArellano2019Towards}.

\subsubsection{Density-based Methods}
The term, density-based method, mainly refers to the selection of samples from the perspective of the set (core set \cite{phillips2016coresets}). The construction of the core set is a representative query strategy. This idea is mainly inspired by the compression idea of the core set dataset and attempts to use the core set to represent the distribution of the feature space of the entire original dataset, thereby reducing the labeling cost of AL. 

FF-Active (Farthest First Active Learning) \cite{Geifman2017DeepActiveLearning} is based on this idea and uses the farthest-first traversal in the space of neural activation over a representation layer to query consecutive points from the pool. It is worth noting here that FF-Active \cite{Geifman2017DeepActiveLearning} and Exploration-P \cite{Yin2017DeepSimilarityBased} resemble the way in which random queries are used in the early stages of AL to enhance AL's exploration ability, which prevents AL from falling into the trap of insufficient sample diversity. Similarly, to solve the sampling bias problem in batch querying, the diversity of batch query samples is increased. 
The Core-set approach \cite{Sener2018ActiveLearningConvolutional} attempts to solve this problem by constructing a core subset. A further attempt was made to solve the k-Center problem \cite{2009Facilitylocationconcepts} by building a core subset so that the model learned on the selected core set will be more competitive than the rest of the data. However, the Core-set approach requires a large distance matrix to be built on the unlabeled dataset, meaning that this search process is computationally expensive; this disadvantage will become more apparent on large-scale unlabeled datasets \cite{Ash2019DeepBatchActive}.

Active Palmprint Recognition \cite{Du2019BuildinganActivePalmprintRecognitionSystem} applies DeepAL to high-dimensional and complex palmprint recognition data. Similar to the core set concept, \cite{Du2019BuildinganActivePalmprintRecognitionSystem} regards AL as a binary classification task. It is expected that the labeled and unlabeled sample sets will have the same data distribution, making the two difficult to distinguish; that is, the goal is to find a labeled core subset with the same distribution as the original dataset. More specifically, due to the heuristic generative model simulation data distribution being difficult to train and unsuitable for high-dimensional and complex data such as palm prints, the author considers whether the sample can be positively distinguished from the unlabeled or labeled dataset with a high degree of confidence. Those samples that can be clearly distinguished are obviously different from the data distribution of the core annotation subset. These samples will then be added to the annotation dataset for the next round of training.
Previous core-set-based methods \cite{Geifman2017DeepActiveLearning, Sener2018ActiveLearningConvolutional} often simply try to query data points as far as possible to cover all points of the data manifold without considering the density, which results in the queried data points overly representing sample points from manifold sparse areas. Similar to \cite{Du2019BuildinganActivePalmprintRecognitionSystem}, DAL (Discriminative Active Learning) \cite{Gissin2018DiscriminativeActiveLearning} also regards AL as a binary classification task and further aims to make the queried labeled dataset indistinguishable from the unlabeled dataset. The key advantage of DAL \cite{Gissin2018DiscriminativeActiveLearning} is that it can sample from the unlabeled dataset in proportion to the data density, without biasing the sample points in the sparse popular domain. Moreover, the method proposed by DAL \cite{Gissin2018DiscriminativeActiveLearning} is not limited to classification tasks, which are conceptually easy to transfer to other new tasks.

In addition to the corresponding query strategy, some researchers have also considered the impact of batch query size on query performance. For example, \cite{BatchBALD2019, Zhdanov2019Diverseminibatch, Ash2019DeepBatchActive, Pinsler2019Bayesian} focus primarily on the optimization of query strategies in smaller batches, while \cite{Chitta2019Training} recommended expanding the query scale of AL for large-scale sampling (10k or 500k samples at a time). Moreover, by integrating hundreds of models and reusing intermediate checkpoints, the distributed searching of training data on large-scale labeled datasets can be efficiently realized with a small computational cost. \cite{Chitta2019Training} also proved that the performance of using the entire dataset for training is not the upper limit of performance, as well as that AL based on subsets specifically may yield better performance.

Furthermore, the attributes of the dataset itself also have an important impact on the performance of DeepAL. With this in mind, GA (Gradient Analysis) \cite{Vodrahalli2018Are} assesses the relative importance of image data in common datasets and proposes a general data analysis tool design to facilitate a better understanding of the diversity of training examples in the dataset. GA \cite{Vodrahalli2018Are} finds that not all datasets can be trained on a small sub-sample set because the relative difference of sample importance in some datasets is almost negligible; therefore, it is not advisable to blindly use smaller sub-datasets in the AL context.
In addition, \cite{DBLP:conf/cvpr/BeluchGNK18} finds that compared with the Bayesian deep learning approach (Monte-Carlo dropout \cite{DBLP:conf/icml/GalIG17}) and density-based \cite{Sener2017geometric} methods, ensemble-based AL can effectively offset the imbalance of categories in the dataset during the acquisition process, resulting in more calibration prediction uncertainty, and thus better performance.

In general, density-based methods primarily consider the selection of core subsets from the perspective of data distribution. There are relatively few related research methods, which suggests a new possible direction for sample querying.

\begin{figure}[!tp] 
	\centering 
	\subfloat[Active learning pipeline.] 
 	{\centering 
 		\includegraphics[width = 0.27\textwidth]{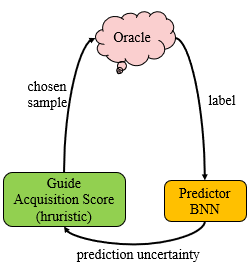}
 		\label{fig:ALp}
 	}\hspace{2mm}
	\subfloat[Reinforced Active Learning (RAL) \cite{Haussmann2019Deep}.] 
 	{\centering 
 		\includegraphics[width = 0.27\textwidth]{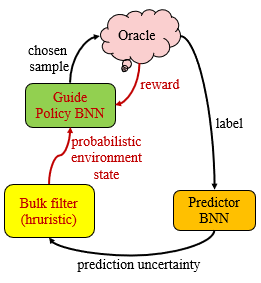}
 		\label{fig:RAL}
 	}\hspace{2mm}
 \subfloat[Deep Reinforcement Active
Learning (DRAL) \cite{Liu2019Deep}.] 
 	{\centering 
 		\includegraphics[width = 0.37\textwidth]{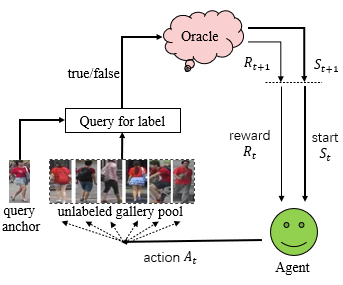}
 		\label{fig:DRAL}
 	}
	\caption{Comparison of standard AL, RAL \cite{Haussmann2019Deep} and DRAL \cite{Liu2019Deep} pipelines.} 
	\label{fig:ALp_RAL} 
\end{figure}

\subsubsection{Automated Design of DeepAL}
DeepAL is composed of two parts: deep learning and active learning. Manually designing these two parts requires a lot of energy and their performance is severely limited by the experience of researchers. Therefore, it has important significance to consider how to automate the design of deep learning models and active learning query strategies in DeepAL.

To this end, \cite{Fang2017Learning} redefines the heuristic AL algorithm as a reinforcement learning problem and introduces a new description through a clear selection strategy. In addition, some researchers have also noted that, in traditional AL workflows, the acquisition function is often regarded as a fixed known prior, and that it will not be known whether this acquisition function is appropriate until the label budget is exhausted. This makes it impossible to flexibly and quickly tune the acquisition function. Accordingly, one good option may be to use reinforcement learning to dynamically tune the acquisition function. 
RAL (Reinforced Active Learning) \cite{Haussmann2019Deep} proposes to use BNN as a learning predictor for acquisition functions. As such, all probability information provided by the BNN predictor will be combined to obtain a comprehensive probability distribution; subsequently, the probability distribution is sent to a BNN probabilistic policy network, which performs reinforcement learning in each labeling round based on the oracle feedback. This feedback will fine-tune the acquisition function, thereby continuously improving its quality. 
DRAL (Deep Reinforcement Active Learning) \cite{Liu2019Deep} adopts a similar idea and designs a deep reinforcement active learning framework for the person Re-ID task. This approach uses the idea of reinforcement learning to dynamically adjust the acquisition function so as to obtain high-quality query samples. Fig.\ref{fig:ALp_RAL} presents a comparison between traditional AL, RAL and DRAL pipelines. 
The pipeline of AL is shown in Fig.\ref{fig:ALp}. The standard AL pipeline usually consists of three parts. The oracle provides a set of labeled data; the predictor (here, BNN) is used to learn these data and provides predictable uncertainty for the guide. The guide is usually a fixed, hard-coded acquisition function that picks the next sample for the oracle to restart the cycle. 
The pipeline of RAL (Reinforced Active Learning) \cite{Haussmann2019Deep} is shown in Fig.\ref{fig:RAL}. RAL replaces the fixed acquisition function with the policy BNN. The policy BNN learns in a probabilistic manner, obtains feedback from the oracle, and learns how to select the next optimal sample point (new parts in red) in a reinforcement learning-based manner. Therefore, RAL can adjust the acquisition function more flexibly to adapt to the existing dataset. 
The pipeline of DRAL (Deep Reinforcement Active Learning) \cite{Liu2019Deep} is shown in Fig.\ref{fig:DRAL}. DRAL utilizes a deep reinforcement active learning framework for the person Re-ID task. For each query anchor (probe), the agent (reinforcement active learner) will select sequential instances from the gallery pool during the active learning process and hand it to the oracle to obtain manual annotation with binary feedback (positive/negative). The state evaluates the similarity relationships between all instances and calculates rewards based on oracle feedback to adjust agent queries.

On the other hand, Active-iNAS (Active Learning with incremental Neural Architecture Search) \cite{Geifman2019Deep} notices that most previous DeepAL methods \cite{Aghdam2019Active, Alahmari2019Automatic, Kwolek2019Breast} assume that a suitable DL model has been designed for the current task, meaning that their primary focus is on how to design an effective query mechanism; however, the existing DL model is not necessarily optimal for the current DeepAL task. Active-iNAS \cite{Geifman2019Deep} accordingly challenges this assumption and uses NAS (neural architecture search) \cite{Ren2020AComprehensiveSurveyofNeuralArchitectureSearchChallengesandSolutions} technology to dynamically search for the most effective model architectures while conducting active learning. 
There is also some work devoted to providing a convenient performance comparison platform for DeepAL; for example, \cite{Munjal2020Towards} discusses and studies the robustness and reproducibility of the DeepAL method in detail, and presents many useful suggestions.

In general, these query strategies are not independent of each other but are rather interrelated. Batch-based BMDAL provides the basis for the update training of AL query samples on the DL model. Although the query strategies in DeepAL are rich and complex, they are largely designed to take the diversity and uncertainty of query batches in BMDAL into account. Previous uncertainty-based methods often ignore the diversity in the batch and can thus be roughly divided into two categories: those that design a mechanism that explicitly encourages batch diversity in the input or learning representation space, and those that directly measure the mutual information (MI) of the entire batch.

\subsection{Data Expansion of Labeled Samples in DeepAL}
\label{sec: Insufficient Data in DeepAL}
AL often requires only a small amount of labeled sample data to realize learning and model updating, while DL requires a large amount of labeled data for effective training. Therefore, the combination of AL and DL requires as much as possible to use the data strategy without consuming too much human resources to achieve DeepAL model training. Most previous DeepAL methods \cite{Zhao2017Deep} often only train on the labeled sample set sampled by the query strategy. However, this ignores the existence of existing unlabeled datasets, meaning that the corresponding data expansion and training strategies are not fully utilized. These strategies help to improve the problem of insufficient labeled data in DeepAL training without adding to the manual labeling costs. Therefore, the study of these strategies is also quite meaningful.

\begin{figure}[!tp] 
	\centering 
	\includegraphics[width = 0.95\textwidth]{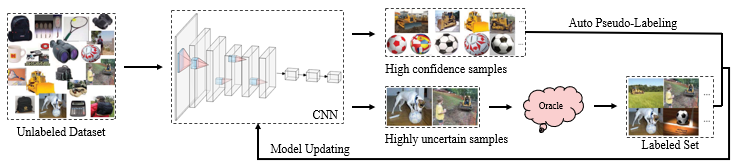}
	\caption{In CEAL \cite{Wang2017CostEffectiveActive}, the overall framework of DeepAL is utilized. CEAL \cite{Wang2017CostEffectiveActive} gradually feeds the samples from the unlabeled dataset to the initialized CNN, after which the CNN classifier outputs two types of samples: a small number of uncertain samples and a large number of samples with high prediction confidence. A small number of uncertain samples are labeled through the oracle, and the CNN classifier is used to automatically assign pseudo-labels to a large number of high-prediction confidence samples. These two types of samples are then used to fine-tune the CNN, and the updated process is repeated.} 
	\label{fig:CEAL} 
\end{figure}

For example, CEAL (Cost-Effective Active Learning) \cite{Wang2017CostEffectiveActive} enriches the training set by assigning pseudo-labels to samples with high confidence in model prediction in addition to the labeled dataset sampled by the query strategy. This expanded training set is then also used in the training of the DL model. This strategy is shown in Fig.\ref{fig:CEAL}.
Another very popular strategy involves performing unsupervised training on labeled and unlabeled datasets and incorporating other strategies to train the entire network structure.
For example, WI-DL \cite{Liu2017Active} notes that full DBN training requires a large number of training samples, and it is impractical to apply DBN to a limited training set in an AL context. Therefore, in order to improve the training efficiency of DBN, WI-DL employs a combination of unsupervised feature learning on all datasets and supervised fine-tuning on labeled datasets.

\begin{figure}[!tp] 
	\centering 
	\subfloat[Generative adversarial active learning (GAAL).] 
 	{\centering 
 		\includegraphics[width = 0.4\textwidth]{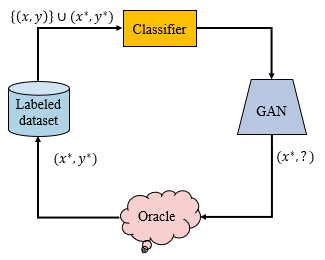}
 		\label{fig:GAAL}
 	}\hspace{3mm}
	\subfloat[Bayesian generative active deep learning (BGADL).] 
 	{\centering 
 		\includegraphics[width = 0.5\textwidth]{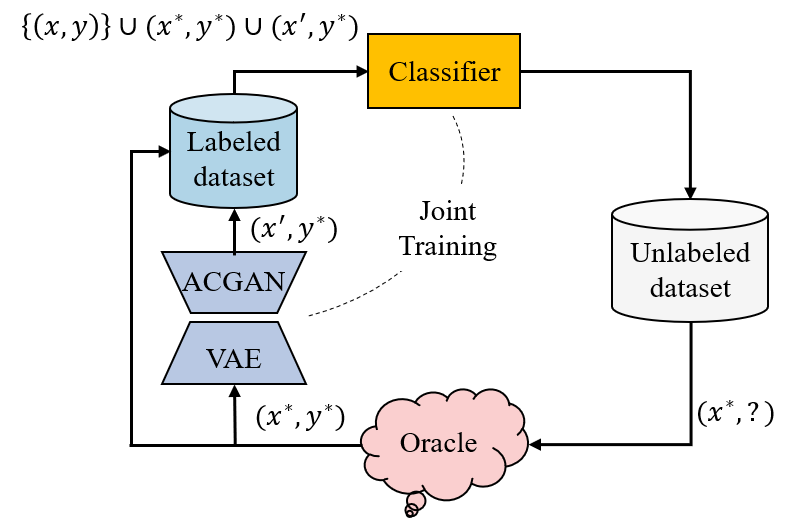}
 		\label{fig:BGADL}
 	}
	\caption{Structure comparison chart of GAAL \cite{Zhu2017Generative} and BGADL \cite{Tran2019BayesianGenerativeActive}. For more details, please see \cite{Tran2019BayesianGenerativeActive}.} 
	\label{fig:GAAL_BGADL} 
\end{figure}

At the same time, some researchers have considered using GAN (Generative Adversarial Networks) for data augmentation. 
For example, GAAL (Generative Adversarial Active Learning) \cite{Zhu2017Generative} introduced the GAN to the AL query method for the first time. GAAL aims to use generative learning to generate samples with more information than the original dataset.
However, random data augmentation does not guarantee that the generated samples will have more information than those contained in the original data, and could thus represent a waste of computing resources. 
Accordingly, BGADL (Bayesian Generative Active Deep Learning) \cite{Tran2019BayesianGenerativeActive} expands the idea of GAAL \cite{Zhu2017Generative} and proposes a Bayesian generative active deep learning method. More specifically, BGADL combines the generative adversarial active learning \cite{Zhu2017Generative}, Bayesian data augmentation \cite{Tran2017Bayesian}, ACGAN (Auxiliary-Classifier Generative Adversarial Networks) \cite{Odena2017Conditional} and VAE (Variational Autoencoder) \cite{Kingma2013Auto} methods, with the aim of generating samples of disagreement regions \cite{Settles2012Active} belonging to different categories. Structure comparison between GAAL and BGADL is presented in Fig.\ref{fig:GAAL_BGADL}.

Subsequently, VAAL \cite{sinha2019variational} and ARAL (Adversarial Representation Active Learning) \cite{Mottaghi2019Adversarial} borrowed from several previous methods \cite{Liu2017Active, Zhu2017Generative, Tran2019BayesianGenerativeActive} not only to train the network using labeled and unlabeled datasets but also to introduce generative adversarial learning into the network architecture for data augmentation purposes, thereby further improving the learning ability of the network.
In more detail, VAAL \cite{sinha2019variational} noticed that the batch-based query strategy based on uncertainty not only readily leads to insufficient sample diversity, but is also highly susceptible to interference from outliers. In addition, density-based methods \cite{Sener2018ActiveLearningConvolutional} are susceptible to $p$-norm limitations when applied to high-dimensional data, resulting in calculation distances that are too concentrated \cite{donoho2000high}. To this end, VAAL \cite{sinha2019variational} proposes to use the adversarial learning representation method to distinguish between the potential spatial coding features of labeled and unlabeled data, thus reducing interference from outliers. 
VAAL \cite{sinha2019variational} also uses labeled and unlabeled data to jointly train a VAE \cite{Kingma2013Auto, Sohn2015Learning} in a semi-supervised manner; the goal here is to deceive the adversarial network \cite{Goodfellow2014Generative} into predicting that all data points come from the labeled pool, in order to solve the problem of distance concentration. VAAL \cite{sinha2019variational} can learn an effective low-dimensional latent representation on a large-scale dataset, and further provides an effective sampling method by jointly learning the representation form and uncertainty.

\begin{figure}[!tp] 
	\centering 
	\includegraphics[width = 0.95\textwidth]{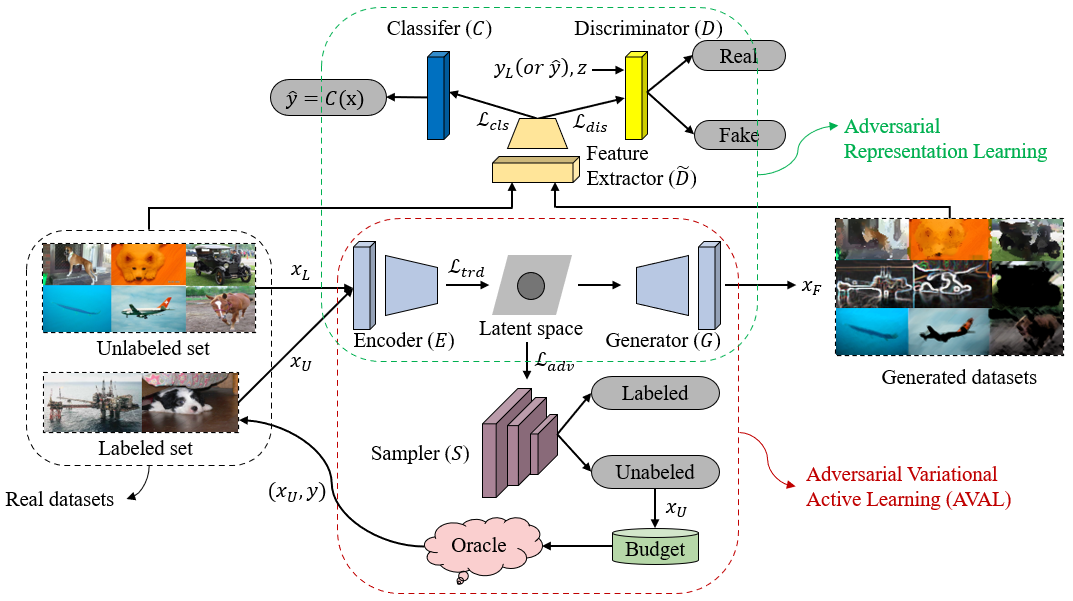}
	\caption{The overall structure of ARAL \cite{Mottaghi2019Adversarial}. ARAL uses not only real datasets (both labeled and unlabeled), but also generated datasets to jointly train the network. The whole network consists of an encoder ($E$), generator ($G$), discriminator ($D$), classifier ($C$) and sampler ($S$), and all parts of the model are trained together.} 
	\label{fig:VAAL_ARAL} 
\end{figure}

Subsequently, ARAL \cite{Mottaghi2019Adversarial} expanded VAAL \cite{sinha2019variational}, aiming to use as few manual annotation samples as possible while still making full use of the existing or generated data information in order to improve the model's learning ability. In addition to using labeled and unlabeled datasets, ARAL \cite{Mottaghi2019Adversarial} also uses samples produced by deep production networks to jointly train the entire model. ARAL \cite{Mottaghi2019Adversarial} comprises both VAAL \cite{sinha2019variational} and adversarial representation learning \cite{Donahue2019Large}. By using VAAL \cite{sinha2019variational} to learn the potential feature representation space of the labeled and unlabeled data, the unlabeled samples with the largest amount of information can be selected accordingly. At the same time, both real and generated data are used to enhance the model's learning ability through confrontational representation learning \cite{Donahue2019Large}. Similarly, TA-VAAL \cite{Kim2020Task} also extends VAAL by using the global data structure from VAAL and local task-related information from the learning loss for sample querying purposes. We present the framework of ARAL \cite{Mottaghi2019Adversarial} in Fig.\ref{fig:VAAL_ARAL}. 

Unlike ARAL \cite{Mottaghi2019Adversarial} and VAAL \cite{sinha2019variational}, which use labeled and unlabeled datasets for adversarial representation learning, SSAL (Semi-Supervised Active Learning) \cite{Simeoni2019Rethinkingdeepactive} implements a new training method. More specifically, SSAL \cite{Simeoni2019Rethinkingdeepactive} uses unsupervised, supervised, and semi-supervised learning methods across AL cycles, and makes full use of existing information for training without increasing the cost of labeling as much as possible. In more detail, the process is as follows: before the AL starts, first use labeled and unlabeled data for unsupervised pretraining. In each AL learning cycle, first, perform supervised training on the labeled dataset, then perform semi-supervised training on all datasets. This represents an attempt to devise a wholly new training method. The author finds that, compared with the difference between the sampling strategies, this model training method yields a surprising performance improvement.

As analyzed above, this kind of exploration of training methods and data utilization skills is also essential; in fact, the resultant performance gains may even exceed those generated by changing the query strategy. Applying these techniques enables the full use of existing data without any associated increase in labeling costs, which helps in resolving the issue of the number of AL query samples being insufficient to support the updating of the DL model.

\subsection{DeepAL Generic Framework}
\label{sec: Common Framework DeepAL}
As mentioned in Section \ref{sec:The necessity and challenge of combining DL and AL}, a processing pipeline inconsistency exists between AL and DL; thus, only fine-tuning the DL model in the AL framework, or simply combining AL and DL to treat them as two separate problems, may cause divergence. For example, \cite{Asghar2016Deep} first conducts offline supervised training of the DL model on two different types of session datasets to grant basic conversational capabilities to the backbone network, then enables the online AL stage to interact with human users, enabling the model to be improved in an open way based on user feedback. AL-DL \cite{Wang2014new} proposes an AL method for DL models with DBNs, while ADN \cite{Zhou2010Active} further proposes an active deep network architecture for sentiment classification. \cite{Stark2015Captcha} proposes an AL algorithm using CNN for captcha recognition. However, generally speaking, the above methods first perform routine supervised training on this depth model on the labeled dataset, then actively sample based on the output of the depth model. There are many similar related works \cite{Shelmanov2019Active, Feng2019Deep} that adopt this split-and-splitting approach that treats the training of AL and deep models as two independent problems and consequently increases the possibility, which the two problems will diverge. Although this method achieved some success at the time, a general framework that closely combines the two tasks of DL and AL would play a vital role in the performance improvement and promotion of DeepAL.

CEAL \cite{Wang2017CostEffectiveActive} is one of the first works to combine AL and DL in order to solve the problem of depth image classification. CEAL \cite{Wang2017CostEffectiveActive} merges deep convolutional neural networks into AL, and consequently proposes a novel DeepAL framework. It sends samples from the unlabeled dataset to the CNN step by step, after which the CNN classifier outputs two types of samples: a small number of uncertain samples and a large number of samples with high prediction confidence. A small number of uncertain samples are labeled by the oracle, and the CNN classifier is used to automatically assign pseudo-labels to a large number of high-prediction-confidence samples. Then, these two types of samples are used to fine-tune the CNN and the update process is repeated. In Fig.\ref{fig:CEAL}, we present the overall framework of CEAL. Moreover, HDAL (Heuristic Deep Active Learning) \cite{Li2017Face} uses a similar framework for face recognition tasks: it combines AL with a deep CNN model to integrate feature learning and AL query model training.

In addition, Fig.\ref{fig:DeepAL} illustrates a widespread general framework for DeepAL tasks. Related works include \cite{Yang2017Suggestive, He2019Towards, Du2019BuildinganActivePalmprintRecognitionSystem, Zhao2020Deeply, Lv2020Deep} , among others. More specifically, \cite{Yang2017Suggestive} proposes a framework that uses an FCN (Fully Convolutional Network) \cite{Long2015Fully} and AL to solve the medical image segmentation problem using a small number of annotations. It first trains FCN on a small number of labeled datasets, then extracts the features of the unlabeled datasets through FCN, using these features to estimate the uncertainty and similarity of unlabeled samples. This strategy, which is similar to that described in Section \ref{sec: Hybrid query strategy}, helps to select highly uncertain and diverse samples to be added to the labeled dataset in order to start the next stage of training. 
Active Palmprint Recognition \cite{Du2019BuildinganActivePalmprintRecognitionSystem} proposes a similar DeepAL framework as that for the palmprint recognition task. The difference is that inspired by domain adaptation \cite{Bendavid2010theory}, Active Palmprint Recognition \cite{Du2019BuildinganActivePalmprintRecognitionSystem} regards AL as a binary classification task: it is expected that the labeled and unlabeled sample sets have the same data distribution, making the two difficult to distinguish. Supervision training can be performed directly on a small number of labeled datasets, which reduces the burden associated with labeling. 
\cite{Lv2020Deep} proposes a DeepAL framework for defect detection. This approach performs uncertainty sampling based on the output features of the detection model to generate a list of candidate samples for annotation. In order to further take the diversity of defect categories in the samples into account, \cite{Lv2020Deep} designs an average margin method to control the sampling ratio of each defect category.

\begin{figure}[!tp] 
	\centering 
	\includegraphics[width = 0.95\textwidth]{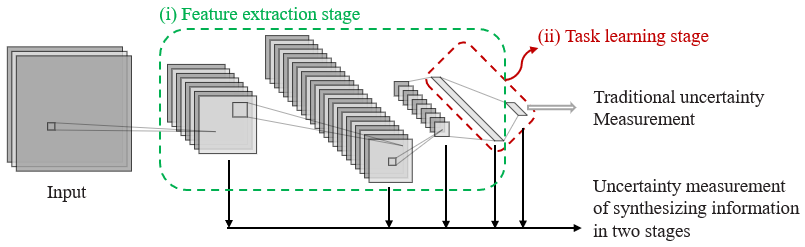}
	\caption{Taking a common CNN as an example, this figure presents a comparison between the traditional uncertainty measurement method \cite{Yang2017Suggestive, Du2019BuildinganActivePalmprintRecognitionSystem, Lv2020Deep} and the uncertainty measurement method of synthesizing information in two stages \cite{He2019Towards, Yoo2019LearningLossActive, Zhao2020Deeply} (i.e., the feature extraction stage and task learning stage).} 
	\label{fig:AL-MV} 
\end{figure}

\begin{figure}[!tp] 
	\centering 
	\includegraphics[width = 0.95\textwidth]{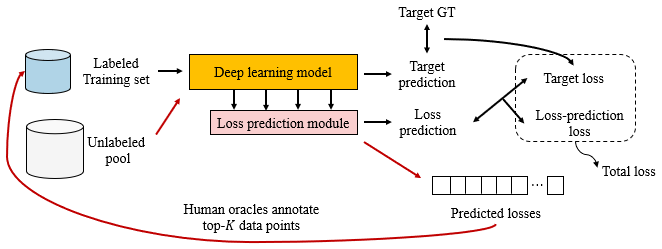}
	\caption{The overall framework of LLAL \cite{Yoo2019LearningLossActive}. The black line represents the stage of training model parameters, optimizing the overall loss composed of target loss and loss-prediction loss. The red line represents the sample query phase of AL. The output of the multiple hidden layers of the DL model is used as the input of the loss prediction module, while the top-$K$ unlabeled data points are selected according to the predicted losses and assigned labels by the oracle.} 
	\label{fig:LLAL} 
\end{figure}

Different from the above methods, it is common for the final output of the DL model to be used as the basis for determining the uncertainty or diversity of the sample (Active Palmprint Recognition \cite{Du2019BuildinganActivePalmprintRecognitionSystem} uses the output of the first fully connected layer). \cite{He2019Towards, Yoo2019LearningLossActive, Zhao2020Deeply} also used the output of the DL model's middle hidden layer. As analyzed in Section \ref{sec: Hybrid query strategy} and Section \ref{sec:The necessity and challenge of combining DL and AL}, due to the difference in learning paradigms between the deep and shallow models, the traditional uncertainty-based query strategy cannot be directly applied to the DL model. In addition, unlike the shallow model, the deep model can be regarded as composed of two stages, namely the feature extraction stage and the task learning stage. It is inaccurate to use only the output of the last layer of the DL model as the basis for evaluating the sample prediction uncertainty; this is because the uncertainty of the DL model is in fact composed of the uncertainty of these two stages. A schematic diagram of this concept is presented in Fig.\ref{fig:AL-MV}. 
To this end, AL-MV (Active Learning with Multiple Views) \cite{He2019Towards} treats the features from different hidden layers in the middle of CNN as multiview data, taking the uncertainty of both stages into account, and the AL-MV algorithm is designed to implement adaptive weighting of the uncertainty of each layer, to enable more accurate measurement of the sampling uncertainty.
 LLAL (Learning Loss for Active Learning) \cite{Yoo2019LearningLossActive} also used a similar idea. More specifically, LLAL designs a small parameter module of the loss prediction module to attach to the target network, using the output of multiple hidden layers of the target network as the input of the loss prediction module. The loss prediction module is learned to predict the target loss of the unlabeled dataset, while the top-$K$ strategy is used to select the query samples. LLAL achieves task-agnostic AL framework design at a small parameter cost and further achieves competitive performance on a variety of mainstream visual tasks (namely, image classification, target detection, and human pose estimation).
Similarly, \cite{Zhao2020Deeply} uses a similar strategy to implement a DeepAL framework for finger bone segmentation tasks. \cite{Zhao2020Deeply} uses Deeply Supervised U-Net \cite{Ronneberger2015U} as the segmentation network, then subsequently uses the output of the multilevel segmentation hidden layer and the output of the last layer as the input of AL; this input information is then integrated to form the basis for the evaluation of the sample information size. 
We take LLAL \cite{Yoo2019LearningLossActive} as an example to explicate the overall network structure of this idea in Fig.\ref{fig:LLAL}.

The research on the general framework is highly beneficial to the development and promotion of DeepAL, as this task-independent framework can be conveniently transplanted to other fields. In the current fusion of DL and AL, DL is primarily responsible for feature extraction, while AL is mainly responsible for sample querying; thus, a deeper and tighter fusion will help DeepAL achieve better performance. Of course, this will require additional exploration and effort on the part of researchers. Finally, the challenges of combining DL and AL and related work on the corresponding solutions are summarized in Table \ref{tab: DAL_challenges_solutions}.

\subsection{DeepAL Stopping Strategy}
\label{sec: DeepAL Stopping Strategy}

In addition to querying strategies and training methods, an appropriate stopping strategy has an important impact on DeepAL performance. At present, most DeepALs \cite{Liu2017Active, Maldonado2019Active, Folmsbee2018Active, Budd2019Survey, schroder2020survey} often use the predefined stopping criterion, and when the criterion is satisfied, they stop querying labels from the oracle. These predefined stopping criteria include the maximum number of iterations, the minimum threshold for changing classification accuracy, the minimum number of labeled samples, and the expected accuracy value, etc.

Although these stopping criteria are simple, these predefined stopping criteria are likely to cause DeepAL to fail to achieve optimal performance. This is because the premature termination of AL annotation querying leads to large performance losses in the model, and excessive annotation behavior wastes a lot of annotation budget. Therefore, Stabilizing Predictions (SP) \cite{DBLP:journals/corr/BloodoodgV14b} makes a comprehensive review of AL stopping strategies and proposes an AL stopping strategy based on stability prediction. Specifically, the SP predivides a part of the samples from the unlabeled dataset to form a stop set (the stop set does not need to be labeled), and the SP checks the prediction stability on the stop set in each iteration. When the prediction performance of the model on the stop set stabilizes, the iteration is stopped. A well-trained model often has a stable predictive ability, and SP takes advantage of this feature. The predivided stop set does not require specific labeling information, which avoids additional labeling costs contrary to the purpose of AL. Although SP is a stopping strategy proposed mainly for AL, it also is relevant for DeepAL.

\begin{table}[!tp]
    \centering
    \scriptsize
    \caption{The challenges of combining DL and AL, as well as a summary of related work on the corresponding solutions.}
    
    \begin{tabular}{|l|l|l|c|l|}\toprule
    \hline
         Challenges & Solutions & Foundation & Category& Publications\\\hline
         \multirow{7}{*}{\makecell[l]{Model\\uncertainty\\in Deep\\Learning}} & \multirow{7}{*}{\makecell[l]{Query\\strategy\\optimization}} & \multirow{7}{*}{\makecell[l]{Batch\\Mode\\DeepAL\\(BMDAL)}}& \makecell[l]{Uncertainty-based\\ and Hybrid \\Query Strategies}&\makecell[l]{\cite{Ranganathan2017Deep, Asghar2016Deep, He2019Towards, Ostapuk2019ActiveLink, Yin2017DeepSimilarityBased, Zhdanov2019Diverseminibatch, Liu2017Active, Ash2019DeepBatchActive, Shui2019DeepActiveLearning, Kim2020Task, Ducoffe2018Adversarial}}\\
         \cline{4-5}
         &&&\makecell[l]{Deep Bayesian\\Active Learning\\(DBAL)}& \makecell[l]{\cite{DBLP:conf/icml/GalIG17, Sener2018ActiveLearningConvolutional, Pop2018DeepEnsembleBayesian, BatchBALD2019, Pinsler2019Bayesian, Chitta2018Large, Ostapuk2019ActiveLink, Siddhant2018Deep, Rottmann2018Deep, Yang2018Leveraging}\\\cite{Zeng2018Relevance, Gudur2019Activeharnet, MartinezArellano2019Towards}}\\
         \cline{4-5}
         &&&\makecell[l]{Density-based\\Methods}& \cite{Geifman2017DeepActiveLearning, Yin2017DeepSimilarityBased, Sener2018ActiveLearningConvolutional, Du2019BuildinganActivePalmprintRecognitionSystem, Gissin2018DiscriminativeActiveLearning, Chitta2019Training}\\
         \cline{4-5}
         &&&\makecell[l]{Automated\\Design of DeepAL}& \cite{Fang2017Learning, Haussmann2019Deep, Liu2019Deep, Geifman2019Deep}\\
         \hline
         \makecell[l]{Insufficient\\data for\\labeled samples} & \makecell[l]{Data\\expansion of\\labeled samples}&\multicolumn{2}{c|}{-}&\cite{Wang2017CostEffectiveActive, Liu2017Active, Zhu2017Generative, Tran2019BayesianGenerativeActive, sinha2019variational, Mottaghi2019Adversarial, Kim2020Task, Simeoni2019Rethinkingdeepactive} \\
         \hline
         \makecell[l]{Processing\\pipeline\\inconsistency} & \makecell[l]{Common\\framework\\DeepAL}&\multicolumn{2}{c|}{-}&\makecell[l]{\cite{Asghar2016Deep, Wang2014new, Zhou2010Active, Stark2015Captcha, Shelmanov2019Active, Feng2019Deep, Wang2017CostEffectiveActive, Li2017Face, Yang2017Suggestive, He2019Towards}\\\cite{Du2019BuildinganActivePalmprintRecognitionSystem, Zhao2020Deeply, Lv2020Deep, Yoo2019LearningLossActive, Vodrahalli2018Are}}\\
         \hline
    \end{tabular}
    \label{tab: DAL_challenges_solutions}
\end{table}

\section{Application of DeepAL in fields such as vision and NLP}
\label{sec: Various applications of DeepAL}
Today, DeepAL has been applied to areas including but not limited to visual data processing (such as object detection, semantic segmentation, etc.), NLP (such as machine translation, text classification, semantic analysis, etc.), speech and audio processing, social network analysis, medical image processing, wildlife protection, industrial robotics, and disaster analysis, among other fields. In this section, we provide a systematic and detailed overview of existing DeepAL-related work from an application perspective.
\subsection{Visual Data Processing}
Just as DL is widely used in the computer vision field, the first field in which DeepAL is expected to reach its potential is that of computer vision. In this section, we mainly discuss DeepAL-related research in the field of visual data processing.
\subsubsection{Image classification and recognition}
As with DL, the classification and recognition of images in DeepAL form the basis for research into other vision tasks. One of the most important problems that DeepAL faces in the field of image vision tasks is that of how to efficiently query samples of high-dimensional data (an area in which traditional AL performs poorly) and obtain satisfactory performance at the smallest possible labeling cost.

To solve this problem, CEAL \cite{Wang2017CostEffectiveActive} assigns pseudo-labels to samples with high confidence and adds them to the highly uncertain sample set queried using the uncertainty-based AL method, then uses the expanded training set to train the DeepAL model image classifier.
\cite{Ranganathan2017Deep} first integrated the criteria of AL into the deep belief network and subsequently conducted extensive research on classification tasks on a variety of real uni-modal and multi-modal datasets.
WI-DL \cite{Liu2017Active} uses the DeepAL method to simultaneously consider the two selection criteria of maximizing representativeness and uncertainty on hyperspectral image (HSI) datasets for remote sensing classification tasks. Similarly, \cite{Lin2018Active,Deng2019Active} also studied the classification of HSI. \cite{Lin2018Active} introduces AL to initialize HSI and then performs transfer learning. This work also recommends constructing and connecting higher-level features to source and target HSI data in order to further overcome the cross-domain disparity. \cite{Deng2019Active} proposes a unified deep network combined with active transfer learning, thereby training the HSI classification well while using less labeled training data.

Medical image analysis is also an important application. For example, \cite{Folmsbee2018Active} explores the use of AL rather than random learning to train convolutional neural networks for tissue (e.g., stroma, lymphocytes, tumor, mucosa, keratin pearls, blood, and background/adipose) classification tasks.
\cite{Budd2019Survey} conducted a comprehensive review of DeepAL-related methods in the field of medical image analysis.
As discussed above, since the annotation of medical images requires strong professional knowledge, it is usually both very difficult and very expensive to find well-trained experts willing to perform annotations. In addition, DL has achieved impressive performance on various image feature tasks. Therefore, a large number of works continue to focus on combining DL and AL in order to apply DeepAL to the field of medical image analysis \cite{Du2018Breast, Sayantan2018Classification, Chen2018Cost, Smailagic2018MedAL, Kwolek2019Breast, Scandalea2019Deep, Smailagic2019O, Sadafi2019Multiclass}.
The DeepAL method is also used to classify in situ plankton \cite{Bochinski2018Deep} and perform the automatic counting of cells \cite{Alahmari2019Automatic}.

In addition, DeepAL also has a wide range of applications in our daily life. For example, \cite{Stark2015Captcha} proposes an AL algorithm that uses CNN for verification code recognition. It can use the ability to obtain labeled data for free to avoid human intervention and greatly improve the recognition accuracy when less labeled data is used.
HDAL \cite{Li2017Face} combines the excellent feature extraction capabilities of deep CNN and the ability to save on AL labeling costs to design a heuristic deep active learning framework for face recognition tasks.

\subsubsection{Object detection and semantic segmentation}
Object detection and semantic segmentation have important applications in various fields, including autonomous driving, medical image processing, and wildlife protection. However, these fields are also limited by the higher sample labeling cost. Thus, the lower labeling cost of DeepAL is expected to accelerate the application of the corresponding DL models in certain real-world areas where labeling is more difficult.

\cite{Roy2018Deep} designs a DeepAL framework for object detection, which uses the layered architecture where labeling is more difficult as an example of "query by committee" to select the image set to be queried, while at the same time introducing a similar exploration/exploitation trade-off strategy to \cite{Yin2017DeepSimilarityBased}.
DeepAL is also widely used in natural biological fields and industrial applications. For example, \cite{Norouzzadeh2019deep} uses deep neural networks to quickly transferable and automatically extract information, and further combines transfer learning and AL to design a DeepAL framework for species identification and counting in camera trap images. 
\cite{Kellenberger2019Half} uses unmanned aerial vehicles (UAV) to obtain images for wildlife detection purposes; moreover, to enable this wildlife detector to be reused, \cite{Kellenberger2019Half} uses AL and introduces transfer sampling (TS) to find the corresponding area between the source and target datasets, thereby facilitating the transfer of data to the target domain. 
\cite{Feng2019Deep} proposes a DeepAL framework for deep object detection in autonomous driving to train LiDAR 3D object detectors.
\cite{Lv2020Deep} proposes the adaptation of a widespread DeepAL framework to defect detection in real industries, along with an uncertainty sampling method for use in generating candidate label categories. This work uses the average margin method to set the sampling scale of each defect category and is thus able to obtain the required performance with less labeled data.

In addition, DeepAL also has important applications in the area of medical image segmentation. For example, \cite{Gaur2016Membrane} proposes an AL-based transfer learning mechanism for medical image segmentation, which can effectively improve the image segmentation performance on a limited labeled dataset.
\cite{Yang2017Suggestive} combines FCN and AL to create a DeepAL framework for biological-image segmentation. This work uses the uncertainty and similarity information provided by the FCN to extend the maximum set cover problem, significantly reducing the required labeling workload by pointing out the most effective labeling areas.
DASL (Deep Active Self-paced Learning) \cite{Wang2018Deepa} proposes a deep region-based network, Nodules R-CNN, for pulmonary nodule segmentation tasks. This work generates segmentation masks for use as examples, and at the same time, combines AL and SPL (Self-Paced Learning) \cite{Kumar2010Self} to propose a new deep active self-paced learning strategy that reduces the labeling workload.
\cite{Wang2019Nodule} proposes a Nodule-plus Region-based CNN for pulmonary nodule detection and segmentation in 3D thoracic Computed Tomography (CT). This work combines AL and SPL strategies to create a new deep self-paced active learning (DSAL) strategy, which reduces the annotation workload and makes effective use of unannotated data.
\cite{Zhao2020Deeply} further proposes a new deep-supervised active learning method for finger bone segmentation tasks. This model can be fine-tuned in an iterative and incremental learning manner and uses the output of the intermediate hidden layer as the basis for sample selection. Compared with the complete markup, \cite{Zhao2020Deeply} achieved comparable segmentation results using fewer samples.

\subsubsection{Video processing} 
Compared with the image task that only needs to process information in the spatial dimension, the video task also needs to process the information in the temporal dimension.
This makes the task of annotating the video more expensive, which also means that the need to introduce AL has become more urgent. DeepAL also has broader application scenarios in this field.

For example, \cite{Hussein2016Deep} proposes to use imitation learning to perform navigation tasks. The visual environment and actions taken by the teacher viewed from a first-person perspective are used as the training set. Through training, it is hoped that students will become able to predict and execute corresponding actions in their own environment. When performing tasks, students use deep convolutional neural networks for feature extraction, learn imitation strategies, and further use the AL method to select samples with insufficient confidence, which are added to the training set to update the action strategy. \cite{Hussein2016Deep} significantly improves the initial strategy using fewer samples.
DeActive \cite{Hossain2018DeActive} proposes a DeepAL activity recognition model. Compared with the traditional DL activity recognition model, DeActive requires fewer labeled samples, consumes fewer resources, and achieves high recognition accuracy.
\cite{Wang2018Deep} minimizes the annotation cost of the video-based person Re-ID dataset by integrating AL into the DL framework. Similarly, \cite{Liu2019Deep} proposes a deep reinforcement active learning method for person Re-ID, using oracle feedback to guide the agent (i.e. the model in the reinforcement learning process) in selecting the next uncertainty sample. The agent selection mechanism is continuously optimized through alternately refined reinforcement learning strategies.
\cite{Aghdam2019Active} further proposes an active learning object detection method based on convolutional neural networks for pedestrian target detection in video and static images.

\subsection{Natural Language Processing (NLP)}
NLP has always been a very challenging task. The goals of NLP are to make computers understand complex human language and to help humans deal with various natural language-related tasks. Insufficient data labeling is also a key challenge in the NLP context. Below, we introduce some of the most famous DeepAL methods in the NLP field.

\subsubsection{Machine translation} Machine translation has very important application value, but it usually requires a large number of parallel corpora as a training set. For many low-resource language pairs, building such a corpus requires a very high cost.

For this reason, \cite{DBLP:conf/ialp/ZhangXX18} proposes to use the AL framework to select information source sentences to construct a parallel corpus. It proposes two effective sentence selection methods for AL: selection based on semantic similarity and decoder probability. Compared with traditional methods, the two proposed sentence selection methods show considerable advantages.
\cite{platanios2019competence} proposes a curriculum learning framework related to AL for machine translation tasks. It can decide which training samples to show to the model during different periods of training based on the estimated difficulty of a sample and the current competence of the model. This method not only effectively improves the training efficiency but also obtains a good accuracy improvement. This kind of thinking is also very valuable for DeepAL's sample selection strategy.

\subsubsection{Text classification} Text classification tasks also face the challenge of excessive labeling costs, such as patent classification \cite{DBLP:conf/dl/Larkey99,DBLP:journals/sigir/FallTBK03} and clinical text classification \cite{DBLP:conf/bionlp/PestianBMHJCD07, DBLP:journals/jamia/FigueroaZNGW12, DBLP:journals/jbi/GarlaTB13}. These labeling tasks often need to be completed by experts, and the number of datasets and texts in each document is often very large, which makes it difficult for human experts to complete the corresponding labeling tasks.

\cite{Zhang2016ActiveDiscriminativeText} claims to be the first AL method for text classification with CNNs. \cite{Zhang2016ActiveDiscriminativeText} focuses on selecting those samples that have the greatest impact on the embedding space. It proposes a method for sentence classification that selects instances containing words whose embeddings are likely to be updated with the greatest magnitude, thereby rapidly learning discriminative, task-specific embeddings. They also extend this method to text classification tasks, which outperformed the baseline AL method in sentence and text classification tasks. \cite{DBLP:conf/icvisp/An0H18} also proposes a new DeepAL framework for text classification tasks. It uses RNN as the acquisition function in AL. The method proposed by \cite{DBLP:conf/icvisp/An0H18} can effectively reduce the number of label instances required for deep learning while saving training time without reducing model accuracy. 
\cite{DBLP:conf/emnlp/PrabhuDS19} focuses on the problem of sampling bias in deep active classification and apply active text classification on the large-scale text corpora of \cite{DBLP:conf/nips/ZhangZL15}. These methods generally show better performance than that of the traditional AL-based baseline methods, and more relevant DeepAL-based text classification applications can be found in \cite{schroder2020survey}.

\subsubsection{Semantic analysis}
In this typical NLP task, the aim is to make the computer understand a natural language description.
The relevant application scenarios are numerous and varied, including but not limited to sentiment classification, news identification, etc. 

More specifically, for example, \cite{Zhou2010Active} uses restricted Boltzmann machines (RBM) to construct an active deep network (ADN), then conduct unsupervised training on the labeled and unlabeled datasets. ADN uses a large number of unlabeled datasets to improve the model's generalization ability, and further employs AL in a semi-supervised learning framework, unifying the selection of labeled data and classifiers in a semi-supervised classification framework; this approach obtains competitive results on sentiment classification tasks.
\cite{Bhattacharjee2017Active} proposes a human-computer collaborative learning system for news accuracy detection tasks (that is, identifying misleading and false information in news) that utilizes only a limited number of annotation samples. This system is a deep AL-based model that uses 1-2 orders of magnitude fewer annotation samples than fully supervised learning. Such a reduction in the number of samples greatly accelerates the convergence speed of the model and results in an astonishing 25\% average performance gains in detection performance.

\subsubsection{Information extraction} Information extraction aims to extract and simplify the most important information from large texts, which is an important basis for correlation analysis between different concepts.

\cite{Priya2019Identifying} uses relevant tweets from disaster-stricken areas to extract information that facilitates the identification of infrastructure damage during earthquakes. For this reason, \cite{Priya2019Identifying} combines RNN and GRU-based models with AL, using AL-based methods to pre-train the model so that it will retrieve tweets featuring infrastructure damage in different regions, thereby significantly reducing the manual labeling workload. 
In addition, entity resolution (ER) is the task of recognizing the same real entities with different representations across databases and represents a key step in knowledge base creation and text mining. 
\cite{Shen2017DeepActiveLearning, Shardlow2019text, Chang2019Overcoming} uses the combination of DL and AL to determine how the technical level of NER (Named Entity Recognition) can be improved in the case of a small training set. \cite{Kasai2019Low} developed a DL-based ER method that combines transfer learning and AL to design an architecture that allows for the learning of a model that is transferable from high-resource environments to low-resource environments.
\cite{Maldonado2019Active} proposes a novel ALPNN (Active Learning Policy Neural Network) design to recognize the concepts and relationships in large EEG (electroencephalogram) reports; this approach can help humans extract available clinical knowledge from a large number of such reports.

\subsubsection{Question-answering}
Intelligent question-answering is also a common processing task in the NLP context, and DL has achieved impressive results in these areas. However, the performance of these applications still relies on the availability of massive labeled datasets; AL is expected to bring new hope to this challenge.

The automatic question-answering system has a very wide range of applications in the industry, and DeepAL is also highly valuable in this field. 
For example, \cite{Asghar2016Deep} uses the online AL strategy combined with the DL model to achieve an open domain dialogue by interacting with real users and learning incrementally from user feedback in each round of dialogue.
\cite{Jedoui2019Deep} finds that AL strategies designed for specific tasks (e.g., classification) often have only one correct answer and that these uncertainty-based measurements are often calculated based on the output of the model. Many real-world vision tasks often have multiple correct answers, which leads to the overestimation of uncertainty measures and sometimes even worse performance than random sampling baselines. For this reason, \cite{Jedoui2019Deep} proposes to estimate the uncertainty in the hidden space within the model rather than the uncertainty in the output space of the model in the Visual Question Answer (VQA) generation, thus overcoming the paraphrasing nature of language.

\subsection{Other Applications}
The emergence of DeepAL is exciting, as it is expected to reduce the annotation costs by orders of magnitude while maintaining performance levels. For this reason, DeepAL is also widely used in other fields.

These applications include, but are not limited to, gene expression, robotics, wearable device data analysis, social networking, ECG signal analysis, etc.
For some more specific examples, MLFS (Multi-Level Feature Selection) \cite{Ibrahim2014Multi} combines DL and AL to select genes/miRNAs based on expression profiles and proposes a novel multi-level feature selection method. MLFS also considers the biological relationship between miRNAs and genes and applies this method to miRNA expansion tasks.
Moreover, the failure risk of real-world robots is expensive. \cite{Andersson2017Deep} proposes a risk-aware resampling technique; this approach uses AL together with existing solvers and DL to optimize the robot's trajectory, enabling it to effectively deal with the collision problem in scenes with moving obstacles, and verify the effectiveness of the DeepAL method on a real nano-quadcopter.
\cite{Zhou2019Active} further proposes an active trajectory generation framework for the inverse dynamics model of the robot control algorithm, which enables the systematic design of the information trajectory used to train the DNN inverse dynamics module.

In addition, \cite{Hossain2019Active, Gudur2019Activeharnet} uses sensors installed in wearable devices or mobile terminals to collect user movement information for human activity recognition purposes. \cite{Hossain2019Active} proposes a DeepAL framework for activity recognition with context-aware annotator selection. ActiveHARNet (Active Learning for Human Activity Recognition) \cite{Gudur2019Activeharnet} proposes a resource-efficient deep ensembled model that supports incremental learning and inference on the device, utilizes the approximation in the BNN to represent the uncertainty of the model, and further proves the feasibility of ActiveHARNet deployment and incremental learning on two public datasets.
For its part, DALAUP (Deep Active Learning for Anchor User Prediction) \cite{Cheng2019Deep} designs a DeepAL framework for anchor user prediction in social networks that reduces the cost of annotating anchor users and improves the prediction accuracy.
DeepAL is also using in the classification of electrocardiogram (ECG) signals. For example, \cite{Rahhal2016Deep} proposes an active DL-based ECG signal classification method. \cite{Hanbay2019Deep} proposed an AL-based ECG classification method using eigenvalues and DL. The use of the AL method enables the cost of marking ECG signals by medical experts to be effectively reduced.
Furthermore, the cost of label annotation in the speech and audio fields is also relatively high.
\cite{Abdelwahab2019Active} finds that a model trained on a corpus composed of thousands of recordings collected by a small number of speakers is unable to be generalized to new domains; therefore, \cite{Abdelwahab2019Active} developed a practical scheme that involves using AL to train deep neural networks for speech emotion recognition tasks when label resources are limited.

In general, the current applications of DeepAL are mainly focused on visual image processing tasks, although there are also applications in NLP and other fields. Compared with DL and AL, DeepAL is still in the preliminary stage of research, meaning that the corresponding classic works are relatively few; however, it still has the same broad application scenarios and practical value as DL. In addition, in order to facilitate readers' access to specific applications of DeepAL in related fields, we have classified and summarized all application scenarios and datasets used by survey-related work in Section \ref{sec: Various applications of DeepAL} in detail. The specific information is shown in Table \ref{tab: Application of DeepAL}.

\begin{table}[!tp]
\caption{DeepAL's research examples in Vision, NLP and other fields.}
    \centering
    \scriptsize
    
    \begin{threeparttable}
        \begin{tabular}{|c|c|l|l|c|}\toprule
        \hline
             Field & Task & Publications & Datasets & Scenes\\ \hline
             \multirow{21}{*}{\makecell[c]{Vision}} & \multirow{7}{*}{\makecell[c]{Image \\classification \\ and\\ recognition}} & \makecell[l]{\cite{Wang2017CostEffectiveActive, Ranganathan2017Deep, Stark2015Captcha}} & \makecell[l]{CACD \cite{DBLP:conf/eccv/ChenCH14}, Caltech-256 \cite{griffin2007caltech}, VidTIMIT \cite{sanderson2008biometric},\\ CK \cite{kanade2000comprehensive}, MNIST \cite{lecun1998gradient}, CIFAR 10 \cite{Krizhevsky2009Learning},\\ emoFBVP \cite{ranganathan2016multimodal}, MindReading \cite{el2004mind} \\Cool PHP CAPTCHA \cite{Stark2015Captcha}} & \makecell[c]{Handwritten numbers, \\face, CAPTCHA\\ recognition, etc.}\\\cline{3-5}
             &&\cite{Liu2017Active, Deng2019Active, Lin2018Active} & \makecell[l]{PaviaC, PaviaU, Botswana \cite{Liu2017Active},\\ Salinas Valley, Indian Pines \cite{Deng2019Active}, \\Washington DC Mall, Urban \cite{Lin2018Active}} &  \makecell{Hyperspectral\\ image}\\
             \cline{3-5}
             &&\makecell[l]{\cite{Folmsbee2018Active, Budd2019Survey, Du2018Breast, Sayantan2018Classification}\\\cite{Smailagic2018MedAL, Kwolek2019Breast, Scandalea2019Deep}\\\cite{Smailagic2019O, Sadafi2019Multiclass, Chen2018Cost}}& \makecell[l]{Erie County \cite{Folmsbee2018Active}, EEG \cite{andrzejak2001indications}, \\BreaKHis \cite{DBLP:journals/tbe/SpanholOPH16}, \\ SVEB, SVDB \cite{Sayantan2018Classification}} & Biomedical\\
             \cline{2-5}
             & \multirow{4}{*}{\makecell[c]{\makecell[c]{Object \\detection}}} & \cite{Roy2018Deep}&VOC \cite{everingham2010pascal}, Kitti \cite{geigerwe} & --  \\
             \cline{3-5}
             &&\cite{Norouzzadeh2019deep, Kellenberger2019Half}& \makecell[l]{SS \cite{swanson2015snapshot}, eMML \cite{forrester2013emammal}, NACTI\tnote{1}, CCT\tnote{2}, UAV\tnote{3}} & \makecell{Biodiversity survey}\\
             \cline{3-5}
             && \cite{Feng2019Deep} &KITTI \cite{DBLP:conf/cvpr/GeigerLU12} & Autonomous driving\\
             \cline{3-5}
             &&\cite{Lv2020Deep} & NEU-DET \cite{song2013noise} & Defect detection\\
             \cline{2-5}
             & \makecell{Semantic \\segmentation} &\cite{Gaur2016Membrane, Yang2017Suggestive, Wang2018Deepa, Wang2019Nodule}& \makecell[l]{SPIM \cite{DBLP:journals/bmcbi/DelibaltovGKKNS16}, Confocal\cite{DBLP:conf/miccai/DelibaltovGRVSM13}, LIDC-IDRI \cite{armato2011lung},\\MICCAI, Lymph node \cite{DBLP:conf/bibm/ZhangYYAC16}}& Bio-medical image \\\cline{2-5}
             & \multirow{5}{*}{\makecell[c]{Video \\processing}} &\cite{Hussein2016Deep} & Mash-simulator\tnote{4} & \makecell[c]{Autonomous\\ navigation}\\
             \cline{3-5}
             &&\cite{Hossain2018DeActive}&\makecell[l]{OPPORTUNITY \cite{Hossain2018DeActive}, WISDM \cite{DBLP:journals/sigkdd/KwapiszWM10},\\ SenseBox \cite{DBLP:conf/percom/TaylorHAKRGG17}, Skoda Daphnet \cite{bachlin2009wearable},CASAS \cite{cook2009assessing}}&Smart home\\
             \cline{3-5}
             &&\cite{Wang2018Deep,Aghdam2019Active}&\makecell[l]{PRID \cite{hirzer2011person}, MARS \cite{DBLP:conf/eccv/ZhengBSWSWT16}, BDD100K \cite{DBLP:journals/corr/abs-1805-04687},\\DukeMTMC-VideoReID \cite{DBLP:conf/cvpr/WuLDYO018},\\ CityPersons \cite{DBLP:conf/cvpr/ZhangBS17}, Caltech Pedestrian\cite{DBLP:journals/pami/DollarWSP12}}&Person Re-id\\
             \hline
             \multirow{14}{*}{\makecell[c]{NLP}} & \makecell[c]{Machine\\ translation} & \cite{DBLP:conf/ialp/ZhangXX18,platanios2019competence}&\makecell[l]{OPUS \cite{DBLP:conf/lrec/Tiedemann12}, UNPC \cite{DBLP:conf/lrec/ZiemskiJP16},\\IWSLT, WMT \cite{DBLP:conf/emnlp/PlataniosSNM18}}&\makecell[c]{Ind-En, Ch-En, En-Vi, \\Fr-En, En-De, etc.}\\
             \cline{2-5}
             & \makecell[c]{Text \\classification}&\cite{Zhang2016ActiveDiscriminativeText, DBLP:conf/icvisp/An0H18, schroder2020survey, DBLP:conf/emnlp/PrabhuDS19}&\makecell[l]{CR\tnote{5}, Subj, MR\tnote{6}, MuR\tnote{7}, DR \cite{DBLP:journals/jamia/WallacePSTD14}\\AGN, DBP, AMZP, AMZF, YRF \cite{DBLP:conf/nips/ZhangZL15}}&-- \\ \cline{2-5}
             & \multirow{3}{*}{\makecell[c]{Semantic \\analysis}} &\cite{Zhou2010Active} &MOV \cite{PangLV02}, BOO, DVDs, ELE, KIT \cite{BlitzerDP07, DasguptaN09}&\makecell[c]{Sentiment \\classification}\\
             \cline{3-5}
             &&\cite{Bhattacharjee2017Active}&\makecell[l]{KDnugget’s Fake News\tnote{8}, \\Harvard Dataverse \cite{kwon2017rumor}, Liar \cite{DBLP:conf/acl/Wang17}}&\makecell[c]{News veracity \\detection}\\
             \cline{2-5}
             &\multirow{5}{*}{\makecell[c]{Information \\extraction}}&\cite{Priya2019Identifying}&Italy, Iran-Iraq, Mexico earthquake dataset & Disaster assessment \\
             \cline{3-5}
             &&\cite{Maldonado2019Active}&Temple University Hospital\tnote{10}& \makecell[c]{Electroencephalography\\(EEG) reports}\\\cline{3-5}
             &&\cite{Shen2017DeepActiveLearning, Shardlow2019text, Chang2019Overcoming, Kasai2019Low}&\makecell[l]{CoNLL \cite{DBLP:conf/conll/SangM03}, NCBI \cite{DBLP:journals/jbi/DoganLL14}, MedMentions \cite{DBLP:conf/acl/McCallumVVMR18},\\OntoNotes \cite{DBLP:conf/conll/PradhanMXNBUZZ13}, DBLP, FZ, AG \cite{DBLP:conf/sigmod/MudgalLRDPKDAR18}, Cora \cite{DBLP:journals/pvldb/WangLYF11}} &\makecell[c]{Named entity \\recognition (NER)}\\
             \cline{2-5}
             &\multirow{2}{*}{\makecell[c]{Question\\answering}}&\cite{Asghar2016Deep}&CMDC \cite{Danescu-Niculescu-Mizil11}, JabberWacky’s chatlogs\tnote{9} &Dialogue generation\\
             \cline{3-5}
             &&\cite{Jedoui2019Deep}&Visual Genome \cite{KrishnaZGJHKCKL17}, VQA \cite{AntolALMBZP15}&\makecell[c]{Visual question answer (VQA)}\\
             \hline
             \multirow{7}{*}{Other} &\multirow{7}{*}{--}&\cite{Ibrahim2014Multi} &BC, HCC, Lung&Gene expression\\\cline{3-5}
             &&\cite{Andersson2017Deep, Zhou2019Active}&EATG \cite{Zhou2019Active}, Crazyflie 2.0\tnote{11} &Robotics\\\cline{3-5}
             &&\cite{Hossain2019Active, Gudur2019Activeharnet}&HHAR \cite{stisen2015smart}, NWFD \cite{DBLP:journals/sensors/MauldinCMNR18}& Smart device\\\cline{3-5}
             &&\cite{Cheng2019Deep}&Foursquare, Twitter \cite{DBLP:conf/cikm/KongZY13}& Social network\\\cline{3-5}
             &&\cite{Rahhal2016Deep, Hanbay2019Deep}&MIT-BIH \cite{mark1982annotated}, INCART, SVDB \cite{Rahhal2016Deep}& \makecell[c]{Electrocardiogram (ECG)\\signal classification}\\\cline{3-5}
             &&\cite{Abdelwahab2019Active}&MSP-Podcast \cite{lotfian2017building} & \makecell[c]{Speech emotion recognition}\\\hline
        \end{tabular}
        \begin{tablenotes}
           \footnotesize
           \item[1] http://lila.science/datasets/nacti
           \item[2] http://lila.science/datasets/caltech-camera-traps
           \item[3] http://kuzikus-namibia.de/xe\_index.html
           \item[4] https://github.com/idiap/mash-simulator 
           \item[5] www.cs.uic.edu/liub/FBS/sentiment-analysis.html
           \item[6] Subj and MR datasets are available at: http://www.cs.cornell.edu/people/pabo/movie-review-data/
           \item[7] http://www.cs.jhu.edu/˜mdredze/datasets/sentiment/
           \item[8] https://github.com/GeorgeMcIntire/fake\_real\_news\_dataset
           \item[9] http://www.jabberwacky.com/j2conversations. JabberWacky is an in-browser, open-domain, retrieval-based bot.
           \item[10] https://www.isip.piconepress.com/projects/tuh\_eeg/
           \item[11] https://www.bitcraze.io/
           \item[--] Non-specific application scenarios
        \end{tablenotes}
    \end{threeparttable}
    \label{tab: Application of DeepAL}
\end{table}

\section{Discussion and future directions}
\label{sec: Discussion and future directions}
DeepAL combines the common advantages of DL and AL: it inherits not only DL's ability to process high-dimensional image data and conduct automatic feature extraction but also AL's potential to effectively reduce annotation costs. DeepAL, therefore, has fascinating potential especially in areas where labels require high levels of expertise and are difficult to obtain.

Most recent work reveals that DeepAL has been successful in many common tasks. DeepAL has attracted the interest of a large number of researchers by reducing the cost of annotation and its ability to implement the powerful feature extraction capabilities of DL; consequently, the related research work is also extremely rich. However, there are still a large number of unanswered questions on this subject.
As \cite{Munjal2020Towards} discovered, the results reported on the random sampling baseline (RSB) differ significantly between different studies. For example, under the same settings, using 20\% of the label data of CIFAR 10, the RSB performance reported by \cite{Yoo2019LearningLossActive} is 13\% higher than that in \cite{Tran2019BayesianGenerativeActive}. Secondly, the same DeepAL method may yield different results in different studies. For example, using 40\% of the label data of CIFAR 100 \cite{Krizhevsky2009Learning} and VGG16 \cite{Simonyan2015VeryDeepConvolutionalNetworksforLargeScaleImageRecognition} as the extraction network, the reported results of \cite{Sener2018ActiveLearningConvolutional} and \cite{sinha2019variational} differ by 8\%. Furthermore, the latest DeepAL research also exhibits some inconsistencies. For example, \cite{Sener2018ActiveLearningConvolutional} and \cite{Ducoffe2018Adversarial} point out that diversity-based methods have always been better than uncertainty-based methods, and that uncertainty-based methods perform worse than RSB; however, the latest research of \cite{Yoo2019LearningLossActive} shows that this is not the case.

Compared with AL's strategic selection of high-value samples, RSB has been regarded as a strong baseline \cite{Yoo2019LearningLossActive, Sener2018ActiveLearningConvolutional}. However, the above problems reveal an urgent need to design a general performance evaluation platform for DeepAL work, as well as to determine a unified high-performance RSB. Secondly, the reproducibility of different DeepAL methods is also an important issue. The highly reproducible DeepAL method helps to evaluate the performance of different DALs. A common evaluation platform should be used for experiments under consistent settings, and snapshots of experimental settings should be shared. In addition, multiple repetitive experiments with different initializations under the same experimental conditions should be implemented, as this could effectively avoid misleading conclusions caused by experimental setup problems. Researchers should pay sufficient attention to these inconsistent studies to enable them to clarify the principles involved. On the other hand, adequate ablation experiments and transfer experiments are also necessary. The former will make it easier for us to determine which improvements bring about performance gains, while the latter can help to ensure that the AL selection strategy does indeed enable the indiscriminate selection of high-value samples for the dataset.

The current research directions regarding DeepAL methods focus primarily on the improvement of AL selection strategies, the optimization of training methods, and the improvement of task-independent models. 
As noted in Section \ref{sec: Query Strategy Optimization in DeepAL}, the improvement of AL selection strategy is currently centered around taking into account the query strategy based on uncertainty and diversity explicitly or implicitly. Moreover, hybrid selection strategies are increasingly favored by researchers. 
Moreover, the optimization of training methods mainly focuses on labeled datasets, unlabeled datasets, or the use of methods such as GAN to expand data, as well as the hybrid training method of unsupervised, semi-supervised, and supervised learning across the AL cycle. This training method promises to deliver even more performance improvements than are thought to be achievable through changes to the selection strategy. In fact, this makes up for the issues of the DL model requiring a large number of labeled training samples and the AL selecting a limited number of labeled samples. In addition, the use of unlabeled or generated datasets is also conducive to making full use of existing information without adding to the annotation costs. Furthermore, the incremental training method is also an important research direction. From a computing resource perspective, it is unacceptable to train a deep model from scratch in each cycle. While simple incremental training will cause the deviation of model parameters, the huge potential savings on resources are quite attractive. Although related research remains quite scarce, this is still a very promising research direction. 

Task independence is also an important research direction, as it helps to make DeepAL models more directly and widely extensible to other tasks. However, the related research remains insufficient, and the corresponding DeepAL methods tend to focus only on the uncertainty-based selection method. Because DL itself is easier to integrate with the uncertainty-based AL selection strategy, we believe that uncertainty-based methods will continue to dominate research directions not related to these tasks in the future. On the other hand, it may also be advisable to explicitly take the diversity-based selection strategy into account; of course, this will also give rise to great challenges. 
In addition, it should be pointed out that blindly pursuing the idea of training models on smaller subsets would be unwise, as the relative difference in sample importance in some datasets with a large variety of content and a large number of samples can almost be ignored.

There is no conflict between the above-mentioned improvement directions; thus, a mixed improvement strategy is an important development direction for the future. In general, DeepAL research has significant practical application value in terms of both labeling costs and application scenarios; however, DeepAL research remains in its infancy at present, and there is still a long way to go in the future.

\section{Summary and conclusions}
\label{sec: Summary and conclusions}
For the first time, the necessity and challenges of combining traditional active learning and deep learning have been comprehensively analyzed and summarized. In response to these challenges, we analyze and compare existing work from three perspectives: query strategy optimization, labeled sample data expansion, and model generality. In addition, we also summarize the stopping strategy of DeepAL. Then, we review the related work of DeepAL from the perspective of the application. Finally, we conduct a comprehensive discussion on the future direction of DeepAL. As far as we know, this is the first comprehensive and systematic review in the field of deep active learning.

\begin{acks}
This work was partially supported by the NSFC under Grant (No.61972315 and No.62072372) and the Shaanxi Science and Technology Innovation Team Support Project under grant agreement (No.2018TD-026) and the Australian Research Council Discovery Early Career Researcher Award (No.DE190100626).
\end{acks}

\bibliographystyle{ACM-Reference-Format}
\normalem
\bibliography{DeepAL}

\appendix

\end{document}